\documentclass{article}

\usepackage{microtype}
\usepackage{graphicx}
\usepackage{subfigure}
\usepackage{booktabs} 

\usepackage{hyperref}

\usepackage[accepted]{icml2025}

\usepackage{amsmath}
\usepackage{amssymb}
\usepackage{mathtools}
\usepackage{amsthm}

\usepackage{enumitem}

\usepackage[capitalize,noabbrev]{cleveref}

\theoremstyle{plain}

\theoremstyle{definition}

\theoremstyle{remark}

\usepackage[textsize=tiny]{todonotes}

\icmltitlerunning{Machine Learning meets Algebraic Combinatoric}

\begin{document}

\twocolumn[
\icmltitle{Machine Learning meets Algebraic Combinatorics: A Suite of Datasets Capturing Research-level Conjecturing Ability in Pure Mathematics}



\icmlsetsymbol{equal}{*}

\begin{icmlauthorlist}
\icmlauthor{Herman Chau}{equal,uw}
\icmlauthor{Helen Jenne}{equal,pnnl}
\icmlauthor{Davis Brown}{equal,pnnl,upenn}
\icmlauthor{Jesse He}{ucsd,pnnl}
\icmlauthor{Mark Raugas}{pnnl}
\icmlauthor{Sara Billey}{uw}
\icmlauthor{Henry Kvinge}{pnnl,uw}
\end{icmlauthorlist}

\icmlaffiliation{uw}{University of Washington}
\icmlaffiliation{pnnl}{Pacific Northwest National Laboratory}
\icmlaffiliation{upenn}{University of Pennsylvania}
\icmlaffiliation{ucsd}{University of California, San Diego}

\icmlcorrespondingauthor{Herman Chau}{hchau@uw.edu}
\icmlcorrespondingauthor{Henry Kvinge}{henry.kvinge@pnnl.gov}


\vskip 0.3in
]



\printAffiliationsAndNotice{\icmlEqualContribution} 

\begin{abstract}
With recent dramatic increases in AI system capabilities, there has been growing interest in utilizing machine learning for reasoning-heavy, quantitative tasks, particularly mathematics. While there are many resources capturing mathematics at the high-school, undergraduate, and graduate level, there are far fewer resources available that align with the level of difficulty and open endedness encountered by professional mathematicians working on open problems. To address this, we introduce a new collection of datasets, the \emph{Algebraic Combinatorics Dataset Repository (ACD Repo)}, representing either foundational results or open problems in algebraic combinatorics, a subfield of mathematics that studies discrete structures arising from abstract algebra. Further differentiating our dataset collection is the fact that it aims at the conjecturing process. Each dataset includes an open-ended research-level question and a large collection of examples (up to 10M in some cases) from which conjectures should be generated. We describe all nine datasets, the different ways machine learning models can be applied to them (e.g., training with narrow models followed by interpretability analysis or program synthesis with LLMs), and discuss some of the challenges involved in designing datasets like these.
\end{abstract}

\section{Introduction}

Modern approaches to machine learning (ML) have been shown to be capable of extracting sophisticated patterns from large and complex datasets. At the same time, there is now increasing evidence that frontier AI systems are capable of performing tasks requiring high-level reasoning capabilities. These trends have led to excitement around the use of machine learning in mathematics. Much of this research explores the use of LLMs and related models to aid in proof writing and mathematical formalization \cite{song2024towards,yang2024leandojo}. While this is an essential part of the mathematician’s workflow, there is also a need for machine assisted conjecture generation using (what we call) `raw' mathematical data. Before identifying a claim that they want to try to prove, a mathematician needs to work through many examples to build intuition and better understand their object of study. For example, when trying to better understand the coefficients of a particular family of polynomials (e.g., Kazhdan-Lusztig polynomials in Section \ref{sect:datasets}), a mathematician may search through countless examples, looking for patterns or other features of interest that may be the basis of future theorems.

Existing applications of machine learning to raw mathematics data tend to fall into several clusters. The first are toy problems (for which we already know many solutions), which are used by the interpretability and science of deep learning communities as a stand-in for more complicated real-world tasks \cite{zhong2024clock, nanda2023progress,liu2023seeing,lee2023teaching}. Another group uses reinforcement learning methods to search for counterexamples to conjectures \cite{charton2024patternboost,mehrabian2023finding,wagner2021constructions}. There is also a growing body of work coming from the mathematics community where off-the-shelf ML methods are just one of several tools used to make progress on a specific problem \cite{coates2024machine, wagner2021constructions, kazalicki2023ranks,bao2022machine,kazalicki2023ranks,davies2021advancing}. Finally in a few instances, foundation models have begun to be deployed to address specific mathematical questions \cite{romera2024mathematical}. 

While these works either present interesting methodological progress in ML or valuable results in mathematics, none aim to provide a range of datasets accessible to the broader ML community that represent open or equivalently challenging research-level problems. To fill the gap we present the {\emph{Algebraic Combinatorics Dataset Repository (ACD Repo)}}\footnote{Datasets and associated code can be found at \url{https://github.com/pnnl/ML4AlgComb}}, a collection of 9 datasets consisting of many examples along with an associated question(s). Our collection includes both open problems (e.g., the combinatorial interpretation of Schubert polynomial structure constants) and classic problems whose solution is a major result in the field (e.g., a combinatorial method of calculating the character of irreducible symmetric group representations). 

We choose to restrict ourselves to algebraic combinatorics (an area of mathematics that studies discrete structures arising from abstract algebra) because (i) it requires less background theory to understand, making it generally more accessible to a broader range of researchers, (ii) there already exist specialized software libraries (e.g., Sage \cite{sage}) designed to efficiently compute many quantities of interest in algebraic combinatorics, and (iii) by nature of being discrete, the objects of interest in algebraic combinatorics tend to be more amendable to representation on a computer. 

Each dataset has both an open-ended mathematical question and a related ML friendly task associated to it. The idea is that a model that can effectively solve the ML task has probably learned information that could offer insight into the broader mathematical question. For example, the open question may be finding a combinatorial interpretation of Schubert polynomial structure constants (\cref{subsect-schubert}), which are indexed by triples of permutations. In this case the ML task is to predict the structure constant from the three permutations. For each dataset, we provide context and motivation for the problem, the basic statistics of the dataset, as well as performance of some basic off-the-shelf models (including reasoning LLMs in a few cases).

We note that these datasets are not designed to be benchmarks in the traditional sense. High performance in terms of standard metrics such as accuracy may be of little value if one is unable to extract mathematical insight that leads to a fruitful conjecture. In Section \ref{sect-case-studies} we give two examples illustrating how this might be done: (i) by performing a careful interpretability analysis of a performant narrow model \cite{davies2021advancing,he2024machines} or (ii) by using a LLM that can communicate its reasoning via (for example) code \cite{austin2021program,romera2024mathematical}. We hope that these datasets will enable the development of even more effective approaches in the future.

\section{The Cast of Characters: Partitions, Permutations, and Partial Orders}\label{ref-background}

The field of combinatorics studies a broad range of problems in mathematics centered around discrete objects (e.g, partial orders, graphs, permutations, partitions) \cite{ stanley2011enumerative1, stanley2011enumerative2}. Ideas and tools from combinatorics play an essential role in many other fields of mathematics and continue to have a strong impact on computer science and physics. Algebraic combinatorics is a subfield of combinatorics that applies combinatorial methods to problems arising from abstract algebra, particularly representation theory and algebraic geometry. 

\begin{figure}[h!]
    \centering
    \includegraphics[scale=0.15]{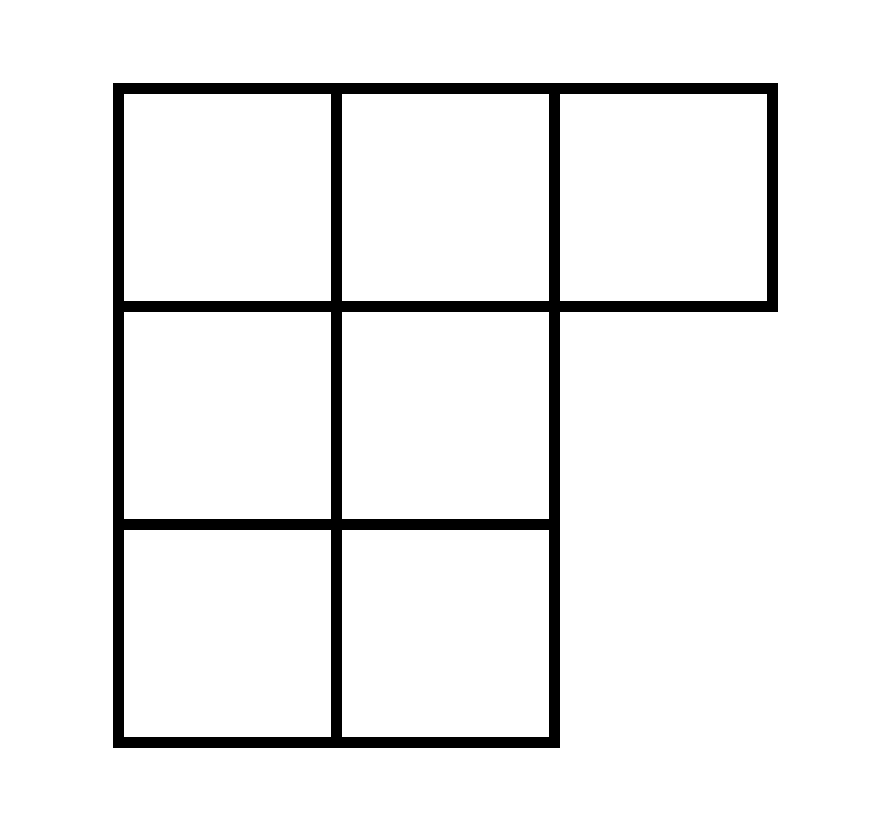}
    \includegraphics[scale=0.15]{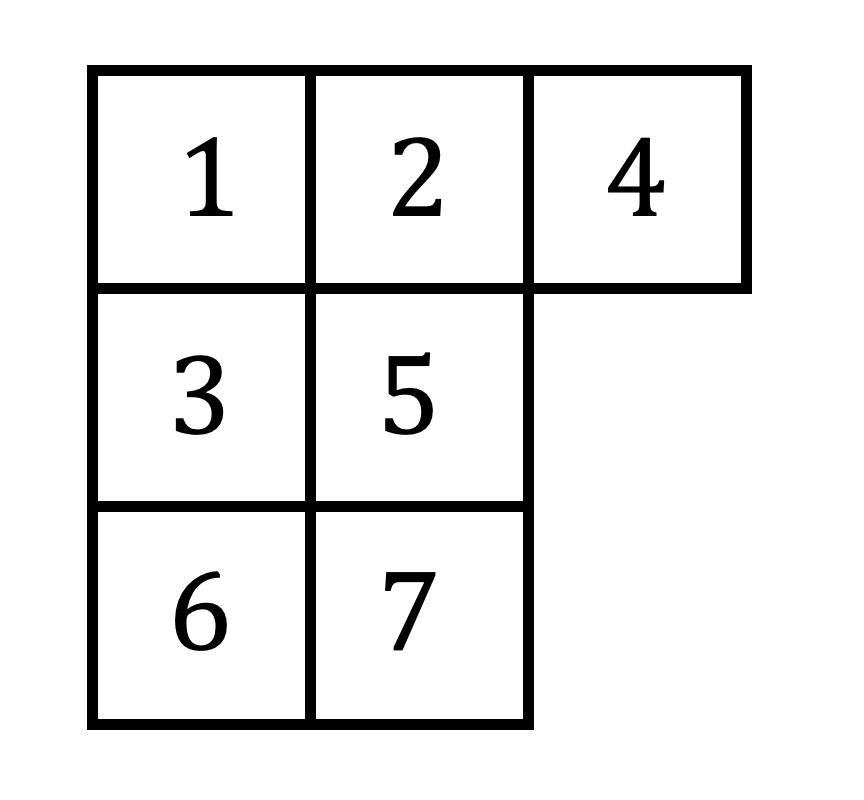}
    \includegraphics[scale=0.15]{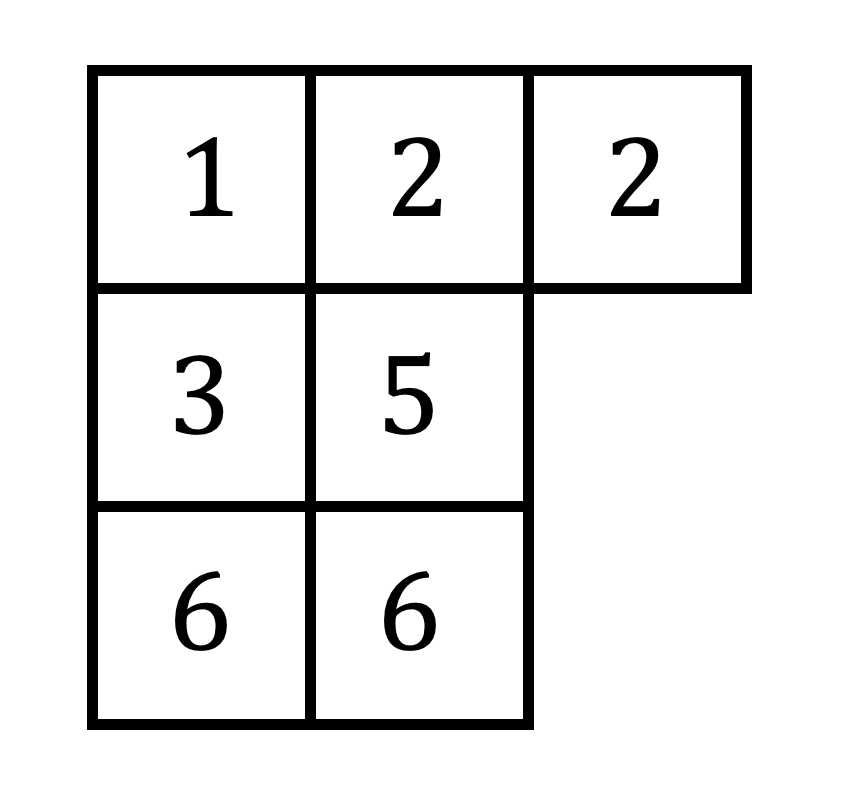}
    \caption{\textbf{(Left)} A Young diagram for the partition $(3,2,2)$. \textbf{(Center)} A standard Young tableau for the partition $(3,2,2)$. \textbf{(Right)} A semistandard Young tableau for the partition $(3,2,2)$.}
    \label{fig:young-examples}
\end{figure}

{\textbf{Partitions:}} We use the word partition in this work to mean an integer partition. An \emph{integer partition of} $n \in \mathbb{N}$ is a sequence of positive integers $(n_1,n_2,\dots,n_k)$ such that $n = n_1 + n_2 + \dots + n_k$ and $n_1 \geq n_2 \geq \dots \geq n_k$. We use the standard notation $\mu \vdash n$ to denote that $\mu$ is a partition of $n$. A partition $(n_1,\dots,n_k)$ is often visualized as a  {\emph{Young diagram}}, with (in English notation) $n_1$ left justified square cells in the first row, $n_2$ left justified square cells in the second row, etc. See Figure \ref{fig:young-examples} (left) for an example of a Young diagram corresponding to the partition $(3,2,2)$.

\textbf{Young tableaux:} Including extra decorations in the cells in a Young diagram can capture universal combinatorics found across representation theory and other fields. A {\emph{Young tableau}} corresponding to a Young diagram $\lambda \vdash n$ is a labeling of the cells of $\lambda$ by an alphabet of symbols. In this work we will consider two types of Young tableaux. A {\emph{standard Young tableau}} corresponding to partition $\lambda \vdash n$ is a labeling of the cells of $\lambda$ by $1,2,\dots,n$ such that the integers strictly increase as one moves down a column or left to right across a row (Figure \ref{fig:young-examples} (center)). The definition of a \emph{semistandard Young tableau} is analogous except that the entries are only assumed to weakly increase as one moves from left to right along a row (see Figure \ref{fig:young-examples} (right)).

\textbf{Permutations:} Permutations are familiar in machine learning from their central role in computer science as well as their relevance to symmetries in many neural networks \cite{entezari2021role,ainsworth2022git,godfrey2022symmetries} and as a symmetry in graph- \cite{keriven2019universal} and set-based problems \cite{zaheer2017deep,lee2019set}. There are many ways to represent a permutation. In this paper we use {\emph{one-line notation}}, which is best illustrated through an example. Suppose $\omega$ is the permutation of the set of elements $\{1,2,3,4\}$ that swaps $1$ and $2$ and $3$ and $4$. Then in one-line notation we write $\omega = 2\;1\;4\;3$. $2$ is in the first position since $1$ is sent to $2$, $1$ is in the second position since $2$ is sent to $1$, $4$ is in the third position since $3$ is sent to $4$, and $3$ is in the fourth position since $4$ is sent to $3$.

Permutations can also be written as sequences of transpositions of adjacent elements. For instance, the permutation $\sigma = 3 \; 1 \; 2$ can be formed by first swapping 2 and 3 and then the newly adjacent 1 and 3: $1 \; 2 \; 3 \rightarrow 1 \; 3 \; 2 \rightarrow 3 \; 1 \; 2$. If we denote a transposition of the $i$th and ($i+1$)st element as $s_{i}$ and read from right to left (as is the convention) then $\sigma$ can be written as $s_1s_2$. A sequence of adjacent transpositions $s_{i_1}s_{i_2}\dots s_{i_k}$ corresponding to a permutation $\sigma$ is called a {\emph{reduced word}} if there is no other representation that uses fewer than $k$ adjacent transpositions to represent $\sigma$. Two reduced words are considered {\emph{commutation equivalent}} if one can be obtained from another by swaps of the form $s_is_j \to s_js_i$ where $|i-j| > 1$. Finally, a {\emph{$3412$ pattern}} is a quadruple $(a_i,a_j,a_k,a_\ell)$ such that $i < j < k < \ell$ but $a_k < a_\ell < a_i < a_j$. Such patterns have deep connections to algebra and geometry \cite{billey1998pattern}.

In the discussion above we implicitly think of permutations of $n$ as bijective functions from $\{1,2,\dots,n\} \rightarrow \{1,2,\dots,n\}$. Using this perspective, one can define the composition of two permutations. The symmetric group, denoted $S_n$, is defined as the group of permutations on $n$ elements using composition as the group operation. The sequence of transpositions $s_1s_2$ from the previous paragraph gave an example of the composition of two permutations. 

{\textbf{Posets:}} A partially ordered set (poset) is a set $P$ of objects equipped with a binary relation, typically denoted ``$\leq$'', that is reflexive, antisymmetric, and transitive. This means that for all elements $a, b, c \in P$: (1) $a \leq a$, (2) if $a \leq b$ and $b \leq a$, then $b=a$, and (3) if $a \leq b$ and $b \leq c$, then $a \leq c$. Unlike total orders which are more familiar (e.g., $\mathbb{Z}$), in a partial order some pairs of elements may be incomparable. An example of a partially ordered set is the set of all subsets of $\{1, 2, 3, 4\}$, ordered by inclusion. This is a partial order and not a total order because $\{1, 2\}$ is not comparable to $\{2, 3\}$ or to $\{2, 3, 4\}$, for example. In a poset, \emph{$y$ covers $x$} if $y$ is greater than $x$ with respect to the ordering, and for any $z$ such that $x \leq z \leq y$, either $z = x$ or $z = y$. In this example, $\{1, 2, 4\}$ covers $\{1, 2\}$, $\{2, 4\},$ and $\{1, 4\}$, but not $\{1\}, \{2\},$ or $\{4\}$.

\section{Related Work}

\textbf{AI for Mathematics:} There is a growing body of work that uses machine learning based methods to assist in mathematics research. While many of these focus more on the proof-creation part of the mathematician's workflow \cite{song2024towards,yang2024leandojo, azerbayev2023llemma}, there are also many that look at the raw mathematical data. This includes the search for counterexamples in graph theory \cite{ wagner2021constructions}, the search for connections between different knot invariants \cite{davies2021advancing}, the classification of $\mathbb{Q}$-Fano varieties \cite{coates2024machine}, and Clifford invariants of ADE Coxeter elements \cite{chen2024machine}. Unlike these works which aim to shed light on specific problems, this paper's goal is to introduce datasets so that both the expert and non-expert can explore the use of machine learning for research-level mathematics problems.

\textbf{Neural Algorithmic Reasoning:} This field explores the application of machine learning methods to problems where traditional symbolic algorithms are more commonly applied \cite{gradient}. Like mathematics, applying machine learning to algorithmic data provides a setting where arbitrarily large amounts of data can be generated. Further, as with mathematics, working with algorithmic data allows us to manipulate certain aspects of the problem (e.g., complexity) in ways not possible in noisy real-world data. The differences between our work and the primary neural algorithmic reasoning benchmark \cite{velivckovic2022clrs}, are substantial. Most notably, the focus of these datasets is not on scientific discovery like the ACD Repo. 
\section{Dataset Descriptions}
\label{sect:datasets}

Further details including dataset statistics, additional problem context, and the method used for generating the datasets can be found in Appendix \ref{appendix:dataset_details}. 

To assess the naive ``hardness'' of each task (not taking into account the actual interpretability of a trained model, which in most cases is the real work), we provide some initial baselines for off-the-shelf models: both narrow (Tables \ref{tab:baselines_classification}, \ref{tab:baselines_regression}, and \ref{tab:baselines_kl}) and LLMs (\cref{tab:baselines_llms_mheight} and \cref{tab:baselines_llms_schubert}). The details of training and testing can be found in \cref{appendix:baseline_hyperparameters}.


\begin{table*}\centering
\begin{tabular}{@{}lcccc@{}}\toprule
\footnotesize{Dataset} & \footnotesize{Logistic regression} & \footnotesize{MLP}  & \footnotesize{Transformer} & \footnotesize{Guessing largest class} \\
\midrule
\scriptsize{Lattice paths} \\
\hspace{10pt}\scriptsize{$n=10$} & \scriptsize{66.2\%} & \scriptsize{90.6\% $\pm$ 0.8\%} & \scriptsize{65.3\% $\pm$ 0.0\%} & \scriptsize{66.2\%} \\
\hspace{10pt}\scriptsize{$n=11$} & \scriptsize{66.3\%} & \scriptsize{95.8\% $\pm$ 0.3\%} & \scriptsize{69.4\% $\pm$ 6.0\%} & \scriptsize{66.3\%} \\
\hspace{10pt}\scriptsize{$n=12$} & \scriptsize{66.5\%} & \scriptsize{98.6 \% $\pm$ 0.1\%} & \scriptsize{86.2\% $\pm$ 14.2\%} & \scriptsize{66.5\%} \\

\scriptsize{Weaving patterns} \\
\hspace{10pt}\scriptsize{$n=6$} & \scriptsize{70.4\%} & \scriptsize{86.1 \% $\pm$ 0.2\%} & \scriptsize{85.9\% $\pm$ 2.3\%} & \scriptsize{63.3\%} \\
\hspace{10pt}\scriptsize{$n=7$} & \scriptsize{85.8\%} & \scriptsize{99.3 \% $\pm$ 0.2\%} & \scriptsize{99.9 \% $\pm$ 0.4\%} & \scriptsize{85.0\%} \\
\scriptsize{Cluster algebra quivers} & \scriptsize{40.3 \%} & \scriptsize{86.5 \% $\pm$ 1.9\%} & \scriptsize{92.9\%$\pm$ 0.5\%} & \scriptsize{17.7\%}\\
\scriptsize{Grassmanian cluster algebras}  \\
\hspace{10pt}\scriptsize{$n=6$} & \scriptsize{65.7\%} & \scriptsize{99.3 \% $\pm$ 0.1\%} & \scriptsize{99.5 \%$\pm$ 0.1\%} & \scriptsize{50.0\%}\\

\scriptsize{Schubert polynomials}  \\
\hspace{10pt}\scriptsize{$n=4$} & \scriptsize{88.8\%} & \scriptsize{93.1 \% $\pm$ 2.6\%} & \scriptsize{94.6\% $\pm$ 1.0\%} & \scriptsize{52.3\%}\\
\hspace{10pt}\scriptsize{$n=5$} & \scriptsize{90.6\%} & \scriptsize{97.5 \% $\pm$ 0.2\%} & \scriptsize{96.2\% $\pm$ 1.1\%} & \scriptsize{49.9\%}\\
\hspace{10pt}\scriptsize{$n=6$} & \scriptsize{89.7\%} & \scriptsize{99.8 \% $\pm$ 0.0\%} & \scriptsize{91.3\% $\pm$ 8.0\%} & \scriptsize{50.1\%}\\
\scriptsize{mHeight}  \\
\hspace{10pt}\scriptsize{$n=8$} & \scriptsize{91.4\%} & \scriptsize{99.4 \% $\pm$ 0.3\%} & \scriptsize{99.7\% $\pm$ 0.4\%}  & \scriptsize{91.4\%}\\
\hspace{10pt}\scriptsize{$n=9$} & \scriptsize{93.2\%} & \scriptsize{99.8 \% $\pm$ 0.6\%} & \scriptsize{99.9\% $\pm$ 0.4\%} & \scriptsize{93.2\%}\\
\hspace{10pt}\scriptsize{$n=10$} & \scriptsize{94.2\%} & \scriptsize{99.9 \% $\pm$ 0.0\%} & \scriptsize{99.9\% $\pm$ 0.6\%} & \scriptsize{94.2\%}\\
\bottomrule
\end{tabular}
\caption{\label{tab:baselines_classification}Off-the-shelf model accuracy on classification datasets. Results are averaged over three random weight initializations with 95\% confidence intervals after a hyperparameter search outlined in Section \ref{appendix:baseline_hyperparameters}}
\vspace{-.3cm}
\end{table*}
\subsection{Computing Characters of Irreducible Symmetric Group Representations (Foundational Result)}

One way to understand the algebraic structure of permutations (symmetric groups, $S_n$) is through their representation theory \cite{sagan2013symmetric}, which converts abstract algebraic questions into linear algebra questions that are often easier to solve. 
A {\emph{representation}} of group $G$ on vector space $V$, is a map $\phi:G \rightarrow GL(V)$ which converts elements of $G$ to invertible linear maps from $V$ to $V$ which respect the compositional structure of the group. A basic result in representation theory says that all representations of a finite group can be decomposed into atomic subspace building blocks called \emph{irreducible representations}. Amazingly, irreducible representations are themselves uniquely determined by the value of the traces, $\text{Tr}(\phi(g))$, where $g$ ranges over subsets of $G$ called conjugacy classes. These values are called {\emph{characters}}. 

The representation theory of symmetric groups has rich combinatorial interpretations. Both the irreducible representations and the conjugacy classes of $S_n$ are indexed by partitions of $n$ and thus the characters of irreducible representations of $S_n$ are indexed by a pair of partitions of $n$. For $\lambda,\mu \vdash n$ we write $\chi^\lambda_\mu$. This combinatorial connection is not superficial; there are algorithms (e.g., the Murnaghan-Nakayama rule \cite{stanley2011enumerative2}), which allow calculation of irreducible characters via simple manipulation of the Young diagrams for $\lambda$ and $\mu$ without any reference to more abstract algebraic structure. We provide datasets for $n = 18, 20, 22$.

\begin{center}
    \includegraphics[width=0.99\linewidth]{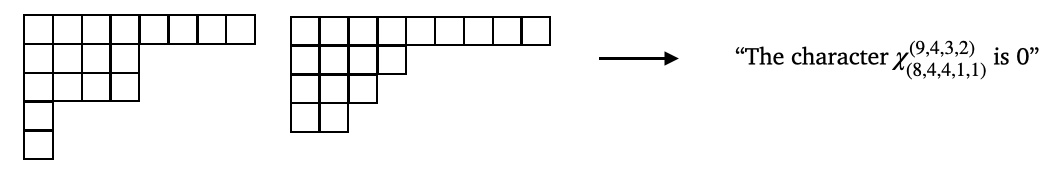}
\end{center}

\textbf{ML task:} Train a model that can take two partitions of $n$, $\lambda$ and $\mu$, and predict the corresponding irreducible symmetric group character $\chi^{\lambda}_{\mu}$. Does this model learn a known algorithm? If it does not, how does the model perform the task? Any novel approaches to computing irreducible symmetric group characters would be of interest to the mathematics community. This is framed as a regression task. 

\textbf{Input/output:} Two integer partitions $\mapsto$ an integer.

\textbf{How hard is it?:} As can be seen in Table \ref{tab:baselines_regression}, narrow models struggle on this task. Some of this may relate to the fact that character distributions are highly concentrated around zero but have long tails (see Figures \ref{fig:symmetric_histogram_18}-\ref{fig:symmetric_histogram_22}). Calculation of characters is also likely to be a hard problem in general \cite{ikenmeyer2024positivity}.

\subsection{The mHeight Function of a Permutation (Key Tool in the Solution to a Recently Solved Conjecture)}

Truly challenging open problems in mathematics often require the development of new mathematical constructions (or even entire new areas of mathematics). This dataset represents a modest example of this. The mHeight function is a statistic associated with a permutation that relates to all $3412$-patterns in the permutation. It was developed and plays a crucial role in the proof by Gaetz and Gao \cite{gaetz-gao} which resolved a long-standing conjecture of Billey and Postnikov \cite{billey-postnikov} about the coefficients on Kazhdan-Lusztig polynomials (Section \ref{subsec-dataset-kl}) The task of predicting the mHeight function represents an interesting opportunity to understand whether a non-trivial intermediate step in an important proof can be learned by machine learning. 

Let $\sigma  = a_1 \ldots a_n \in S_n$ be a permutation containing at least one occurrence of a $3412$ pattern. Let $(a_i,a_j,a_k,a_\ell)$ be a $3412$ pattern so that $i < j < k < \ell$ but $a_k < a_\ell < a_i < a_j$ (see Section \ref{ref-background} for more details). The \emph{height} of $(a_i,a_j,a_k,a_\ell)$ is $a_i - a_\ell$. The \emph{mHeight} of $\sigma$ is then the minimum height over all $3412$ patterns in $\sigma$. We provide datasets for $n = 8,9,10$.

\includegraphics[width=0.99\linewidth]{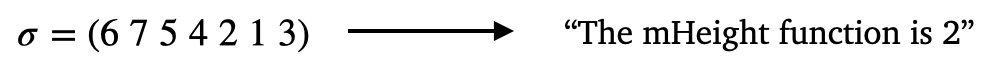}

\textbf{ML task:} Predict the mHeight of a permutation. Since mHeight can take a limited number of values for small $n$, this is framed as a classification problem. 

\textbf{Input/output:} One permutation $\mapsto$ one of several possible integers corresponding to mHeight values.

\textbf{How hard is it?:} This seems to be one of the easier tasks to solve in this collection (at least from the perspective of accuracy alone). Both LLMs ( \cref{tab:baselines_llms_mheight}) and narrow models ( \cref{tab:baselines_classification}) achieve fairly high-accuracy (though note that the dataset is heavily imbalanced for larger $n$). Understanding whether the models actually learn some approximation of mHeight or exploit other correlations remains to be studied.

\subsection{Grassmannian Cluster Algebras and Semistandard Young Tableaux (Open Problem)}

The Grassmann manifold $\text{Gr}(k,n)$ is the set of full-rank $k \times n$ matrices up to equivalence of elementary row operations (equivalently the space whose points are $k$-dimensional subspaces in $\mathbb{R}^n$). Grassmannians are of fundamental geometric importance and are a central tool in a model of quantum field theory known as supersymmetric Yang-Mills theory \cite{golden2014motivic}. 

Among the many algebraic-combinatorial properties of Grassmannians is an algebraic structure on its coordinate ring making it something called a cluster algebra \cite{williams2014cluster}. A recent result of Chang, Duan, Fraser, and Li \cite{chang-duan-fraser-li} parameterize cluster variables of the Grassmannian coordinate ring in terms of equivalence classes of semistandard Young tableaux (SSYT). Not every SSYT indexes a cluster variable and a natural question to ask is which are valid cluster variable indices. A necessary condition is that the tableau is of rectangular shape. We follow the set-up and use the positive examples provided by \cite{cheung2022clustering} who first applied machine learning to this problem, though we choose a different method of sampling the tableaux that do not index cluster variables. The math question is thus to find a concise combinatorial characterization of those SSYT that index a cluster variable. We provide a dataset for $\text{Gr}(3,12)$, restricting to rank $4$. This corresponds to rectangular SSYT of shape $3 \times 4$ with entries drawn from $\{1,2,\dots,12\}$.

\begin{center}
    \includegraphics[width=0.8\linewidth]{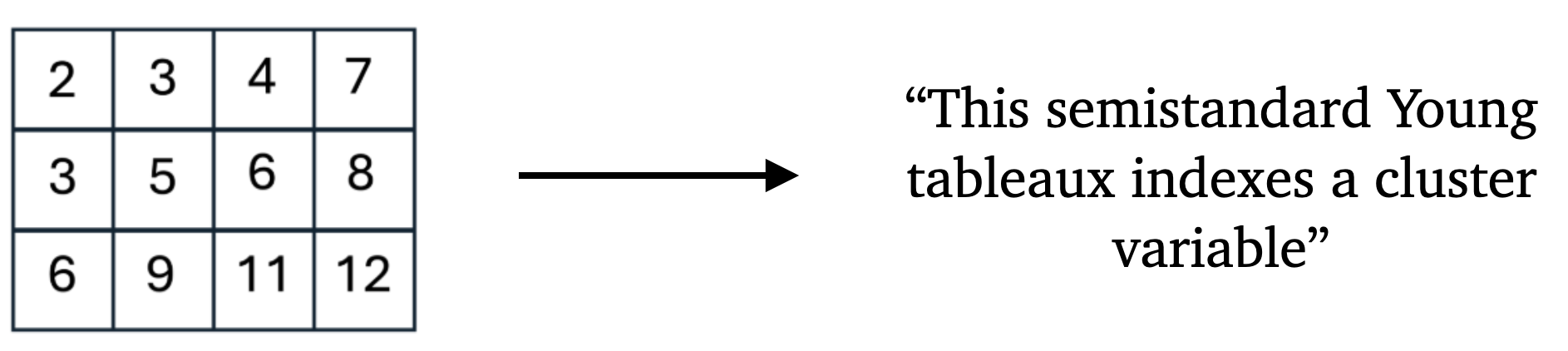}
\end{center}

\textbf{ML task:} Predict whether a Young tableau indexes a cluster variable. This is a binary classification task. 

\textbf{Input/output:} A SSYT of shape $3 \times 4$ with entries drawn from $\{1,2, \dots,12\}$ $\mapsto$ Boolean indicating whether the tableau indexes a cluster variable or not.

\textbf{How hard is it?:} Even naive methods can result in high accuracy (\cref{tab:baselines_classification}), suggesting that there is something interesting here to learn. 
\begin{table}\centering
\begin{tabular}{@{}lccc@{}}\toprule
\footnotesize{Dataset} & \footnotesize{o1 Mini} & \footnotesize{GPT-4o Mini}  & \footnotesize{GPT-4o} \\
\midrule
\scriptsize{In-context learning}  \\
\hspace{10pt}\scriptsize{$n=10$} & \scriptsize{$95.4\%$ / 0.13} & \scriptsize{$89.5\%$ / 0.04} &  \scriptsize{$95.5\%$ / 0.18}\\
\hspace{10pt}\scriptsize{$n=11$} & \scriptsize{$97.0\%$ / 0.22} & \scriptsize{$97.0\%$ / 0} &  \scriptsize{$97.0\%$ / 0.08}\\
\scriptsize{Program synthesis}  \\
\hspace{10pt}\scriptsize{$n=10$} & \scriptsize{$94.2\%$ / 0} & \scriptsize{$94.2\%$ / 0} &  \scriptsize{$94.2\%$ / 0}\\
\hspace{10pt}\scriptsize{$n=11$} & \scriptsize{$95.5\%$ / 0.01} & \scriptsize{$95.5\%$ / 0 } &  \scriptsize{$95.5\%$ / 0.11}\\
\bottomrule
\end{tabular}
\caption{\label{tab:baselines_llms_mheight}Performance of Claude 3.5 Sonnet, GPT-4o Mini, and GPT-4o solving the mHeight function task via either in-context learning or program synthesis. Hyperparameters for these experiments can be found in Section \ref{appendix:llm_hyperparameters}. The results are reported as accuracy/MCC, where the Matthews Correlation Coefficient (MCC) provides a balanced metric that ranges from -1 (total disagreement) to 1 (perfect prediction), with 0 corresponding to chance-level performance. Note that for $n=10$, naive class-guessing (largest class: “0”) already achieves 94.2\% accuracy (and has a 0 MCC).}
\vspace{-.3cm}
\end{table}

\subsection{The Coefficients of Kazhdan-Lusztig Polynomials (Open Problem)}
\label{subsec-dataset-kl}

Kazhdan-Lusztig (KL) polynomials are integer polynomials in a variable $q$ that (for our purposes) are indexed by a pair of permutations \cite{kazhdan1979representations}. We will write the KL polynomial associated with permutations $x$ and $w$ as $P_{x,w}(q)$. For example, the KL polynomial associated with permutations $x = 1\;4\;3\;2\;7\;6\;5\;10\;9\;8\;11$ and $w = 4\;6\;7\;8\;9\;10\;1\;11\;2\;3\;5$ is
{\tiny{
\begin{equation*}
P_{x,w}(q) = 1 + 16q + 103q^2 + 337q^3 + 566q^4 + 529q^5 + 275q^6 + 66q^7 + 3q^8
\end{equation*}
}}
(this example was computed by \cite{warringtonwebsite}). KL polynomials have deep connections throughout several areas of mathematics. For example, KL polynomials are related to the dimensions of intersection homology in Schubert calculus, the study of the Hecke algebra, and the representation theory of the symmetric group. They can be computed via a recursive formula \cite{kazhdan1979representations}, nevertheless, in many ways they remain mysterious. For instance, there is no known closed formula for the degree of $P_{x,w}(q)$.

Mathematicians would like a better understanding of the coefficients on powers of $q$ in $P_{x,w}(q)$. For example, one question concerns the coefficient on term $q^{\ell(x) - \ell(w)-1/2}$ (where $\ell(x)$ is a statistic called the length of the permutation), which is known as the $\mu$-coefficient \cite{warrington2011equivalence}. Both the constant term and coefficient on $q$ are well-understood. We provide datasets for $n = 5,6,7$ using code from \cite{warringtonwebsite}. 

\begin{center}
    \includegraphics[width=0.8\linewidth]{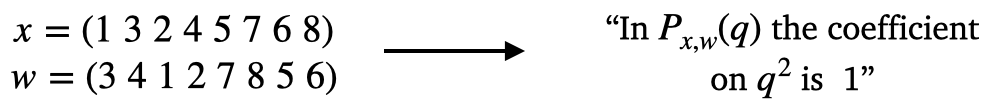}
\end{center}

\textbf{ML Tasks:} Predict one or more coefficients of $P_{x,w}(q)$ given $x$ and $w$. Empirically, for fixed $n$ and power of $q$, coefficients tend to only take a few values, so we frame this as a classification task.

\textbf{Input/output:} Two permutations $\mapsto$ one of several integers.

\textbf{How hard is it?:} We provide both accuracy (\cref{tab:baselines_kl}) and macro F1-scores (\cref{tab:baselines_kl_f1}) for these imbalanced datasets. Narrow models perform well in general.

\subsection{The Robinson-Schensted-Knuth Correspondence (Foundational Result)}

The Robinson-Schensted-Knuth (RSK) algorithm \cite{robinson1938representations, schensted1961longest} gives a bijection between pairs of semistandard Young tableaux of the same shape and matrices with non-negative integer entries. The special case we consider (which is sometimes called the Robinson-Schensted algorithm) restricts to a bijection between pairs of standard Young tableaux and permutations in $S_n$. This correspondence is significant in algebraic combinatorics because it connects two of the most fundamental objects in the field (see \cite{stanley-1984} for additional history and context). 

Given its fundamental importance, it would be interesting to see whether a model learns the RSK algorithm given enough examples of the correspondence. To this end the dataset consists of triples: two standard Young tableaux of size $n$ and their corresponding permutation (obtained via the RSK algorithm). We provide datasets for $n = 8,9$.

\begin{center}
    \includegraphics[width=0.8\linewidth]{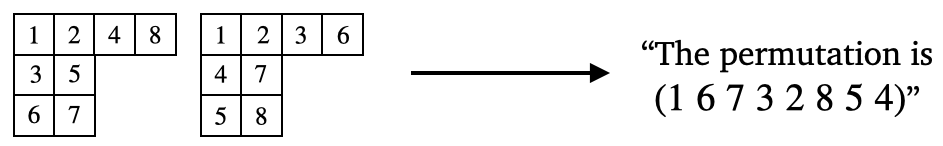}
\end{center}

\textbf{ML task:} Given a pair of standard Young tableaux, predict the corresponding permutation. The task is framed as regression and the permutation target is represented as a vector representing its descent set (we found this to be an easier target for regression than representing the permutation in 1-line notation). 

\textbf{Input/output:} Two standard Young tableaux of the same shape $\mapsto$ a permutation.

\textbf{How hard is it?:} None of the small, narrow models learned this task in our experiments, achieving results similar or worse than simply guessing the mean of all permutations (\cref{tab:baselines_regression}). Like symmetric group character calculation, the RSK algorithm has intricate combinatorial rules that may require larger and more capable models or more elaborate training strategies.

\subsection{Schubert Polynomial Structure Constants (Open Problem)}
\label{subsect-schubert}

Schubert polynomials \cite{bernstein1973, demazure, lascoux1982} are a family of polynomials indexed by permutations of $S_n$. Developed to study the cohomology ring of the flag variety, they have deep connections to algebraic geometry, Lie theory, and representation theory. Despite their geometric origins, Schubert polynomials can be described completely combinatorially \cite{billey1993some, bergeron1993}, making them a well-studied object in algebraic combinatorics. An important open problem in the study of Schubert polynomials is understanding their {\em structure constants}. 
When two Schubert polynomials are multiplied, their product is a linear combination of Schubert polynomials, $\mathfrak{S}_{\alpha} \mathfrak{S}_{\beta} = \sum_{\gamma} c^{\gamma}_{\alpha \beta} \mathfrak{S}_{\gamma}$. The $c^{\gamma}_{\alpha \beta}$ are conjectured to have a combinatorial description or formula (likely related to permutations $\alpha$, $\beta$, and $\gamma$). To give an example of what we mean by combinatorial description, the structure constants of Schur polynomials (a special type of Schubert polynomial) count the number of semistandard tableaux satisfying certain properties.

Each instance in this dataset is a triple of permutations $(\alpha, \beta, \gamma)$, labeled by its coefficient $c^{\gamma}_{\alpha,\beta}$ in the expansion of the product $\mathfrak{S}_{\alpha} \mathfrak{S}_{\beta}$. Not all possible triples of permutations are included; the dataset consists of an approximately equal number of zero and nonzero coefficients. We provide datasets for $n = 4, 5, 6$.

\begin{center}
    \includegraphics[width=0.99\linewidth]{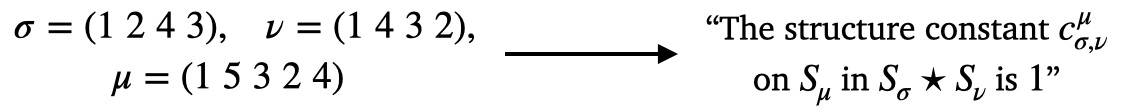}
\end{center}

\textbf{ML task:} Train a model that given three permutations $\alpha, \beta, \gamma$, can predict the associated structure constant $c^{\gamma}_{\alpha,\beta}$. Since these structure constants only take a few values for small permutations, we frame this as a classification task.

\textbf{Input/output:} Three permutations $\mapsto$ one of several integers.

\textbf{How hard is it?:} Both small MLPs and transformers can achieve high-accuracy (\cref{tab:baselines_classification}) as well as LLMs via program synthesis (\cref{tab:baselines_llms_schubert}). Some of the latter is an artifact of how we originally sampled zero-valued structure constants.


\subsection{Partial Orders on Lattice Paths (Open Problem)} 

Consider northeast lattice paths that travel along the edges of a grid from $(0,0)$ to $(a,b)$, only taking steps north and east and never passing through the diagonal $y = \frac{b}{a}x$, where $a$ and $b$ are relatively prime. \cite{schiffler2023perfect} defines two order relations on such paths called the {\em matching ordering} ($\leq_M$) and the {\em Lagrange ordering} ($\leq_L$), motivated by questions in number theory. The matching ordering assigns a number to each lattice path based on the number of perfect matchings of an associated snake graph, while the Lagrange ordering assigns a number to each lattice path equal to the Lagrange number of a certain continued fraction. These numbers each define the respective partial order. Mathematicians would be interested to better understand the relationship between these orders \cite{apruzzese2023}. We provide datasets for lattice paths from zero to $(10,9)$, $(11,10)$, $(12,11)$, and $(13,12)$.  

\begin{center}
    \includegraphics[width=0.8\linewidth]{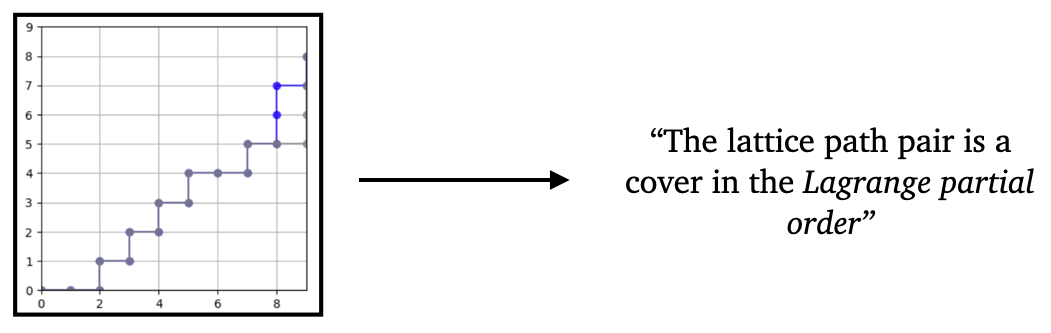}
\end{center}

\textbf{ML task:} Given a pair of lattice paths $(w,w')$, train a model that can predict whether $w'$ covers $w$ (see Section \ref{ref-background}) in either the matching or Lagrange order. 

\textbf{Input/output:} Two lattice paths $\mapsto$ whether the pair is a cover relation in matching or Lagrange order.

\textbf{How hard is it?:} MLPs achieve good performance (\cref{tab:baselines_classification}). On the other hand, we found it challenging to train performant transformers. 
\subsection{Mutation Equivalence of Quivers (Open Problem)}
\label{subsect-quivers}

Quivers and quiver mutations are central to the combinatorial study of cluster algebras, algebraic structures with connections to Poisson Geometry, string theory, and Teichmuller theory. Quivers are directed (multi)graphs, and a quiver mutation is a local transformation centered at a chosen node of the graph that involves adding, deleting and reversing the orientation of specific edges based on a set of combinatorial rules. A fundamental open problem in this area is finding an algorithm that determines whether two quivers are mutation equivalent (one can traverse from one quiver to another by applying mutations). Currently, such algorithms only exist for special cases, including types $A$ \cite{buan2008derived}, $D$ \cite{vatne2010mutation}, and $\tilde{B}$, $\tilde{C}$, and $\tilde{D}$ \cite{henrich2011}. The $\tilde{B}$ and $\tilde{C}$ types correspond to the classes $BD$ and $BB$ in our dataset, respectively. Consistent with Sage we use the naive notation, which specifies a quiver by indicating the two ends of the diagram, which are joined by a path \cite{musiker2011compendium}. To our knowledge, the remaining classes in this dataset ($E$, $DE$, $BE$) lack characterizations. Recent work has explored whether deep learning models can learn to correctly predict if two quivers are mutation equivalent \cite{bao2020quiver}. \cite{he2024machines} utilized an alternative version of this dataset to re-discover known characterization theorems. The dataset consists of adjacency matrices for quivers drawn from 7 different mutation equivalence classes ($A$, $D$, $E$, $DE$, $BE$, $BD$, and $BB$).

\begin{center}
    \includegraphics[width=0.8\linewidth]{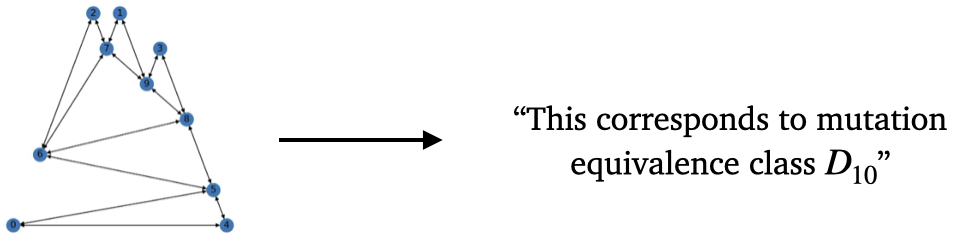}
\end{center}

\textbf{Task:} Find simple rules to determine which of 7 different mutation equivalence classes a quiver belongs to. This is framed as a classification task.

\textbf{Input/output:} An adjacency matrix $\mapsto$ one of 7 different mutation equivalence classes.

\textbf{How hard is it?:} Transformers and MLPs achieve reasonable accuracy on this task (
\cref{tab:baselines_classification}). \cite{he2024machines} was able to train a far more performant model (accuracy $>99\%$) and re-discover several known characterization theorems using a specialized graph neural network architecture.

\subsection{Weaving Patterns (Open Problem)}

Weaving patterns are $n \times n-1$-matrices with $\{1,2,\dots,n\}$-entries introduced by \citet{felsner} to study the number of reduced decompositions of the permutation $\sigma = n \; n-1 \; \ldots 1$ up to commutation equivalence. The number of such objects also counts the number of parallel sorting networks, the number of rhombic tilings of regular polygons, and is connected to the study of the higher Bruhat orders \cite{chau2024enumerating}. 
An $O(n^2)$ algorithm for determining if a given $\{1,2,\dots,n\}$-matrix is a valid weaving pattern exists but gives no additional insight into the structure of weaving patterns and correspondingly the asymptotics of reduced decompositions.

The enumeration of reduced decompositions up to commutation equivalence has been studied by many including Knuth and Stanley. An exact formula is likely out of reach, so asymptotic upper and lower bounds are of great interest. ML models that can detect necessary or sufficient conditions for a matrix to be a valid weaving pattern have the potential to lead to improvements in the upper bound.

Each dataset is a mixture of weaving patterns and non-weaving pattern matrices with $\{1, 2, \ldots, n\}$-entries. We provide datasets for $n = 6,7$.

\begin{center}
    \includegraphics[width=0.8\linewidth]{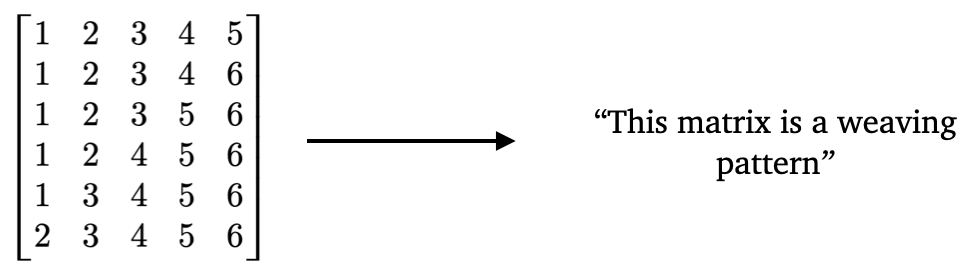}
\end{center}

\textbf{ML task:} Train a model to classify whether a $\{1,2,\dots,n\}$-matrix is a weaving pattern or not. This task is framed as binary classification.

\textbf{Input/output:} A $n \times n-1$ $\{1, 2, \ldots, n\}$-matrix $\mapsto$ binary label.

\textbf{How hard is it?:} Especially in the $n=7$ regime, even small MLPs and transformers achieve high accuracy (\cref{tab:baselines_classification}).

\section{Case Studies of Applying ML to these Datasets}
\label{sect-case-studies}

Mathematical conjecture generation is not a commonly studied task within mainstream ML. We therefore begin this section by providing two examples for how conjectures can be generated using a combination of standard and machine learning-based tools. Our purpose is not to propose novel ML methods nor to describe novel mathematics, but rather to provide the reader with prototype examples of how the ACD repository can be used in practice.

\textbf{Graph Neural Networks and XAI to Better Understand Quiver Mutation:} \cite{davies2021advancing} was one of the first works to show that mathematical conjectures can be made through careful analysis of a deep neural network trained to solve a task around an open problem in mathematics. The basic outline is (1) train a model on the ML friendly task related to the open problem, (2) analyze the resulting performant model to understand how it is making its predictions, and (3) generate conjectures based on this analysis. 

In \cite{he2024machines} we applied this approach to an altered version of the quiver mutation equivalence dataset from the ACD repo (\cref{subsect-quivers}). We showed that we could rediscover a characterization theorem of type $D$ and type $\tilde{D}$ quivers using off-the-shelf deep learning explainability tools applied to a specialized graph neural network architecture. As noted in \cref{subsect-quivers}, concise characterizations that allow one to determine whether one quiver can be transformed into another quiver through a sequence of mutations are of interest to both mathematicians and mathematical physicists and are only known in a handful of cases. 

To do this we trained a new variant of a message passing graph neural network called a DirGINE specifically designed to be able to capture the presence of subgraphs that are important in prediction (several characterization theorems for quivers hinge on identifying the presence or absence of certain subgraph motifs). Having trained a model that achieves high accuracy differentiating between the different mutation equivalence classes included in the dataset, we used latent space clustering to identify distinct cases that needed to be explored and then used a graph neural network explainability technique called PGExplainer \cite{luo2020parameterized} to identify the subgraph motifs that characterized each of these clusters. In the end we were able to re-discover theorems first introduced in \cite{vatne2010mutation} and \cite{henrich2011}.  

\textbf{Program Synthesis for Schubert Polynomials:} The previous example highlighted use of narrow models for conjecture generation. In our second example we describe the use of foundation models for a less labor- but more compute-intensive version of conjecture generation. This work uses the paradigm established by \cite{romera2024mathematical}. The basic idea is that one should structure inference so that an LLM produces code rather than raw predictions since code is intrinsically human interpretable. This is often known as {\emph{program synthesis}}.

We applied this procedure to the Schubert polynomial structure constants dataset. There are various additional steps that one can use in program synthesis to improve code generation (for instance, in \cite{romera2024mathematical} the authors interleave code generation with the application of an evolutionary algorithm). For our experiments we only performed a single round of $100$ generated Python programs with no additional steps. We were surprised to find that (if we provided the proper mathematical context in our prompt) o1-mini and GPT-4o were able to produce Python programs that achieved 100\% accuracy on the test set. To generate the dataset, we subsample from the set of zero valued structure constants because these are far more common than non-zero valued structure constants. Interestingly, closer examination showed that the LLMs that achieved 100\% had reverse engineered our sampling algorithm. This original sampling algorithm essentially involved applying a single transposition to one of the three permutations indexing a non-zero structure constant to obtain a zero structure constant. It turns out that in the cases we applied program synthesis to ($n = 4, 5$), this process introduced a spurious correlation in the dataset\footnote{This issue has since been fixed.}: structure constant $c^{\gamma}_{\alpha,\beta} \neq 0$ if and only if $\ell(\alpha) + \ell(\beta) + \ell(\gamma) \neq 0 \mod 2$ where $\ell$ is a statistic on permutations called the length. We note that the domain relevant notion of permutation length is not explicitly included in the dataset itself or in the prompt we gave the model to initiate program synthesis.

Though this exercise did not provide any useful mathematical insights, it does show that the LLM-based approach to conjecture generation carries the benefit that in some cases the LLM can exploit background tools and concepts (such as the length of a permutation) that need to be learned from scratch by narrow models. More speculatively, it may be that program synthesis tends to elicit such background information to a greater extent than other methods. For comparison, asking these models to solve this task directly using in-context learning, resulted in only 50-60\% accuracy.

{\textbf{Challenges when generating mathematics datasets:}} Generating useful datasets for mathematics problems presents a number of challenges. For instance, imbalance is an issue in several different respects. On the one hand, traditional class imbalance comes up frequently, but there can also be imbalance in terms of how interesting the examples of a dataset are. For a given task, it may be the case that the vast majority of randomly sampled instances are uninteresting because they can be predicted or classified for straightforward reasons. In these cases, individual instances do not capture the mathematics that we care about. When enough data exists, one way to mitigate this situation is to subsample for harder examples. This is what we did for a number of the datasets in the ACD Repo including Weaving Patterns where we imposed some additional constraints on the non-weaving pattern $\{1,2,\dots,n\}$-matrices to make them harder to distinguish from true weaving patterns. The danger then is that the machine learning algorithm just learns this sampling strategy.

The choice of input representation can also have large downstream impacts on how hard it is for a model to learn to solve a task. For example, there are many equivalent ways to represent a permutation. Like other parity prediction tasks \cite{hahn2024sensitive}, prediction of permutation parity is a hard task for transformers when the permutation is presented in one-line notation. Models do substantially better when input permutations are represented via their inversion vector which is more straightforwardly related to permutation parity. We opted to provide all data in formats that are either common in the field or convenient for representation on a computer, leaving it to future researchers to find more optimized representations.

\section{Conclusion and Limitations}
\label{sect:conclusion}

In this paper we introduce the Algebraic Combinatorics Dataset Repository, a collection of research-level mathematics datasets structured for machine learning and designed to accelerate mathematical discovery.  While we believe that these datasets will provide significant value to the ML community, they also have some limitations. Namely, we think we made reasonable design choices around the problems, tasks, and data formats, however, rapid advancement in the field of AI for math means that we can't be certain of this. Nonetheless, we expect AI/ML tools to open new avenues for research. We therefore hope that our datasets will be useful to researchers looking to understand and make progress in this area.






\section*{Acknowledgements}
This research was supported by the Mathematics for Artificial Reasoning in Science (MARS)
initiative at Pacific Northwest National Laboratory. It was conducted under the Laboratory Directed
Research and Development (LDRD) Program at at Pacific Northwest National Laboratory (PNNL), a
multiprogram National Laboratory operated by Battelle Memorial Institute for the U.S. Department
of Energy under Contract DE-AC05-76RL01830.

\bibliographystyle{icml2025}
\bibliography{icml}


\newpage
\appendix
\onecolumn

\begin{table*}\centering
\begin{tabular}{@{}lcccc@{}}\toprule
\footnotesize{Dataset} & \footnotesize{Linear regression} & \footnotesize{MLP}  & \footnotesize{Transformer} & \footnotesize{Guess training label mean} \\
\midrule

\scriptsize{$S_n$ characters}  \\
\hspace{10pt}\scriptsize{$n=18$} & \scriptsize{$1.5920\times 10^{10}$} & \scriptsize{$2.7447\times 10^9  \pm 8.86015\times 10^8$} &  \scriptsize{$2.4913\times10^{10}\pm 1.4350\times10^{7}$} & \scriptsize{$1.5920\times10^{10}$}\\
\hspace{10pt}\scriptsize{$n=20$} & \scriptsize{$4.2007\times 10^{12}$} & \scriptsize{$4.2254\times10^{12} \pm 5.1236\times 10^{11}$} & \scriptsize{$5.3897\times10^{12} \pm 3.6464\times10^{11}$} & \scriptsize{$4.2007\times10^{12}$}\\
\hspace{10pt}\scriptsize{$n=22$} & \scriptsize{$8.0395\times 10^{14}$} & \scriptsize{$1.1192\times10^{14} \pm 4.9321\times10^{12}$} & \scriptsize{$1.3797\times10^{14} \pm 6.2799\times10^{12}$} & \scriptsize{$8.0395\times10^{14}$}\\
\scriptsize{RSK}  \\
\hspace{10pt}\scriptsize{$n=8$} & \scriptsize{0.21} & \scriptsize{0.43  $\pm$ 0.05} &  \scriptsize{1.51 $\pm$ 0.02} & \scriptsize{0.21}\\
\hspace{10pt}\scriptsize{$n=9$} & \scriptsize{0.21} & \scriptsize{0.96  $\pm$ 0.07} & \scriptsize{3.85 $\pm$ 0.09} & \scriptsize{0.21}\\
\bottomrule
\end{tabular}
\caption{\label{tab:baselines_regression}Off-the-shelf model accuracy on regression datasets. Results are averaged over three random weight initializations with 95\% confidence intervals after a hyperparameter search outlined in Section \ref{appendix:baseline_hyperparameters}.}
\vspace{-.3cm}
\end{table*}
\begin{table*}\centering
\begin{tabular}{@{}lccc@{}}\toprule
\footnotesize{Dataset/coefficient} & \footnotesize{MLP}  & \footnotesize{Transformer} & \footnotesize{Guessing largest class} \\
\midrule

\scriptsize{$n = 5$}  \\
\hspace{10pt}\scriptsize{$1$} &  \scriptsize{99.8\% $\pm$ 0.2\%} &  \scriptsize{99.9\% $\pm$ 0.1\%} & \scriptsize{73.7\%}\\
\hspace{10pt}\scriptsize{$q$} &  \scriptsize{99.5\% $\pm$ 0.4\%} & \scriptsize{99.2\% $\pm$ 1.0\%} & \scriptsize{97.0\%}\\
\hspace{10pt}\scriptsize{$q^2$}  & \scriptsize{99.9\% $\pm$ 0.1\%} & \scriptsize{100.0\% $\pm$ 0.0\%} & \scriptsize{99.9\%}\\
\scriptsize{$n = 6$}  \\
\hspace{10pt}\scriptsize{$1$}  & \scriptsize{99.9\% $\pm$ 0.0\%} &  \scriptsize{100.0\% $\pm$ 0.0\%} & \scriptsize{80.9\%}\\
\hspace{10pt}\scriptsize{$q$}  & \scriptsize{99.9\% $\pm$ 0.0\%} & \scriptsize{99.9 $\pm$ 0.0} & \scriptsize{95.8\%}\\
\hspace{10pt}\scriptsize{$q^2$}  & \scriptsize{99.9\% $\pm$ 0.0\%} & \scriptsize{99.9 $\pm$ 0.0} & \scriptsize{99.5\%}\\
\hspace{10pt}\scriptsize{$q^3$}  & \scriptsize{99.9\% $\pm$ 0.0\%} & \scriptsize{99.9 $\pm$ 0.0} & \scriptsize{99.9\%}\\
\bottomrule
\end{tabular}
\caption{\label{tab:baselines_kl}Baseline model classification accuracy for predicting KL polynomial coefficients for $n = 5,6$. Results are averaged over three random weight initializations with 95\% confidence intervals. The MLPs had layer dimension $256$ and depth $4$ and were trained with a learning rate of $0.0005$. The transformers had dimension $256$, depth $6$, $8$ heads, and were trained with a learning rate of $0.0005$.}
\vspace{-.3cm}
\end{table*}

\begin{table*}\centering
\begin{tabular}{@{}lcccc@{}}\toprule
\footnotesize{Dataset/coefficient}  & \footnotesize{MLP}  & \footnotesize{Transformer}  \\
\midrule

\scriptsize{$n = 5$}  \\
\hspace{10pt}\scriptsize{$1$}  & \scriptsize{99.7\% $\pm$ 0.1\%} &  \scriptsize{99.9\% $\pm$ 0.4\%} \\
\hspace{10pt}\scriptsize{$q$}  & \scriptsize{93.9\% $\pm$ 3.7\%} & \scriptsize{92.7\% $\pm$ 7.6\%} \\
\hspace{10pt}\scriptsize{$q^2$} & \scriptsize{50.0\% $\pm$ 0.0\%} & \scriptsize{100.0\% $\pm$ 0.0\%} \\
\scriptsize{$n = 6$}  \\
\hspace{10pt}\scriptsize{$1$}  & \scriptsize{99.9\% $\pm$ 0.0\%} &  \scriptsize{100.0\% $\pm$ 0.0\%} \\
\hspace{10pt}\scriptsize{$q$}  & \scriptsize{99.0\% $\pm$ 1.5\%} & \scriptsize{98.0 $\pm$ 3.7} \\
\hspace{10pt}\scriptsize{$q^2$} & \scriptsize{97.4\% $\pm$ 5.2\%} & \scriptsize{86.2 $\pm$ 47.7} \\
\hspace{10pt}\scriptsize{$q^3$} & \scriptsize{87.9\% $\pm$ 4.5\%} & \scriptsize{88.3 $\pm$ 17.1} \\
\bottomrule
\end{tabular}
\caption{\label{tab:baselines_kl_f1}Baseline model classification macro F1-scores for predicting KL polynomial coefficients for $n = 5,6$. Results are averaged over three random weight initializations with 95\% confidence intervals. The MLPs had layer dimension $256$ and depth $4$ and were trained with a learning rate of $0.0005$. The transformers had dimension $256$, depth $6$, $8$ heads, and were trained with a learning rate of $0.0005$.}
\vspace{-.3cm}
\end{table*}
\begin{table*}\centering
\begin{tabular}{@{}lccc@{}}\toprule
\footnotesize{Dataset} & \footnotesize{Claude 3.5 Sonnet} & \footnotesize{GPT-4o Mini}  & \footnotesize{GPT-4o} \\
\midrule

\scriptsize{In-context learning}  \\
\hspace{10pt}\scriptsize{$n=3$} & \scriptsize{$76.4\%$} & \scriptsize{$64.7\%$} &  \scriptsize{$58.8\%$}\\
\hspace{10pt}\scriptsize{$n=4$} & \scriptsize{$59.5\%$} & \scriptsize{$53.5\%$} &  \scriptsize{$57.0\%$}\\
\hspace{10pt}\scriptsize{$n=5$} & \scriptsize{$58.5\%$} & \scriptsize{$51.5\%$} & \scriptsize{$57.0\%$}\\
\scriptsize{Program synthesis}  \\
\hspace{10pt}\scriptsize{$n=3$} & \scriptsize{$94.1\%$} & \scriptsize{$82.4\%$} &  \scriptsize{$100.0\%$}\\
\hspace{10pt}\scriptsize{$n=4$} & \scriptsize{$65.0\%$} & \scriptsize{$100.0\%$} &  \scriptsize{$100.0\%$}\\
\hspace{10pt}\scriptsize{$n=5$} & \scriptsize{$99.8\%$} & \scriptsize{$64.2\%$} & \scriptsize{$64.7\%$}\\
\bottomrule
\end{tabular}
\caption{\label{tab:baselines_llms_schubert}Success of Claude 3.5 Sonnet, GPT-4o Mini, and GPT-4o solving the Schubert polynomial structure constant task via either in-context learning or program synthesis. Hyperparameters for these experiments can be found in Section \ref{appendix:llm_hyperparameters}. As described in \cref{sect-case-studies}, models that achieved $100\%$ had effectively learned the strategy by which we subsampled zero-valued structure constants. Our subsampling strategy has since been refined.}
\vspace{-.3cm}
\end{table*}

\appendix

\begin{figure}[h!]
    \centering
    \includegraphics[scale=0.30]{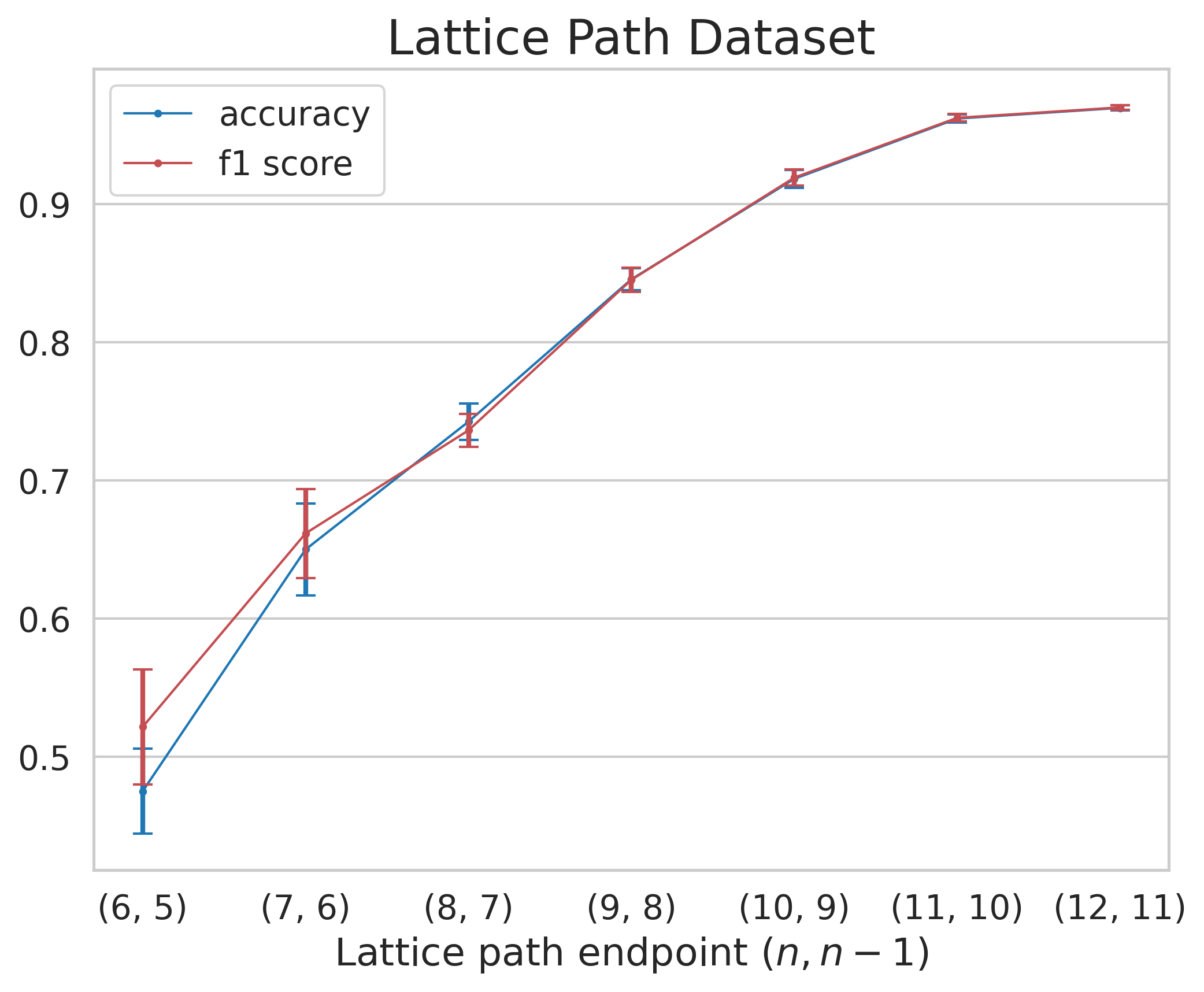}
    \includegraphics[scale=0.30]{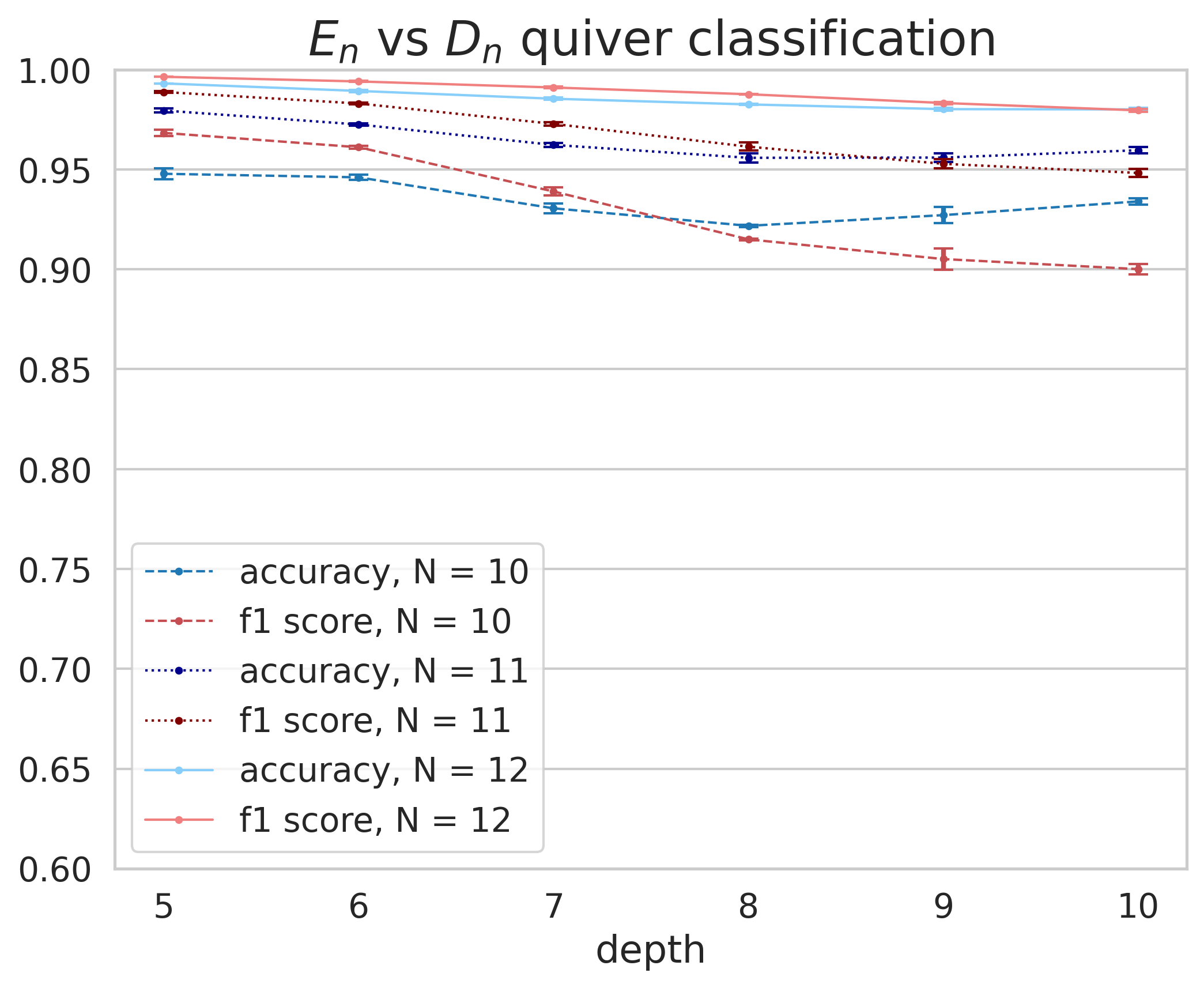}
    \includegraphics[scale=0.30]{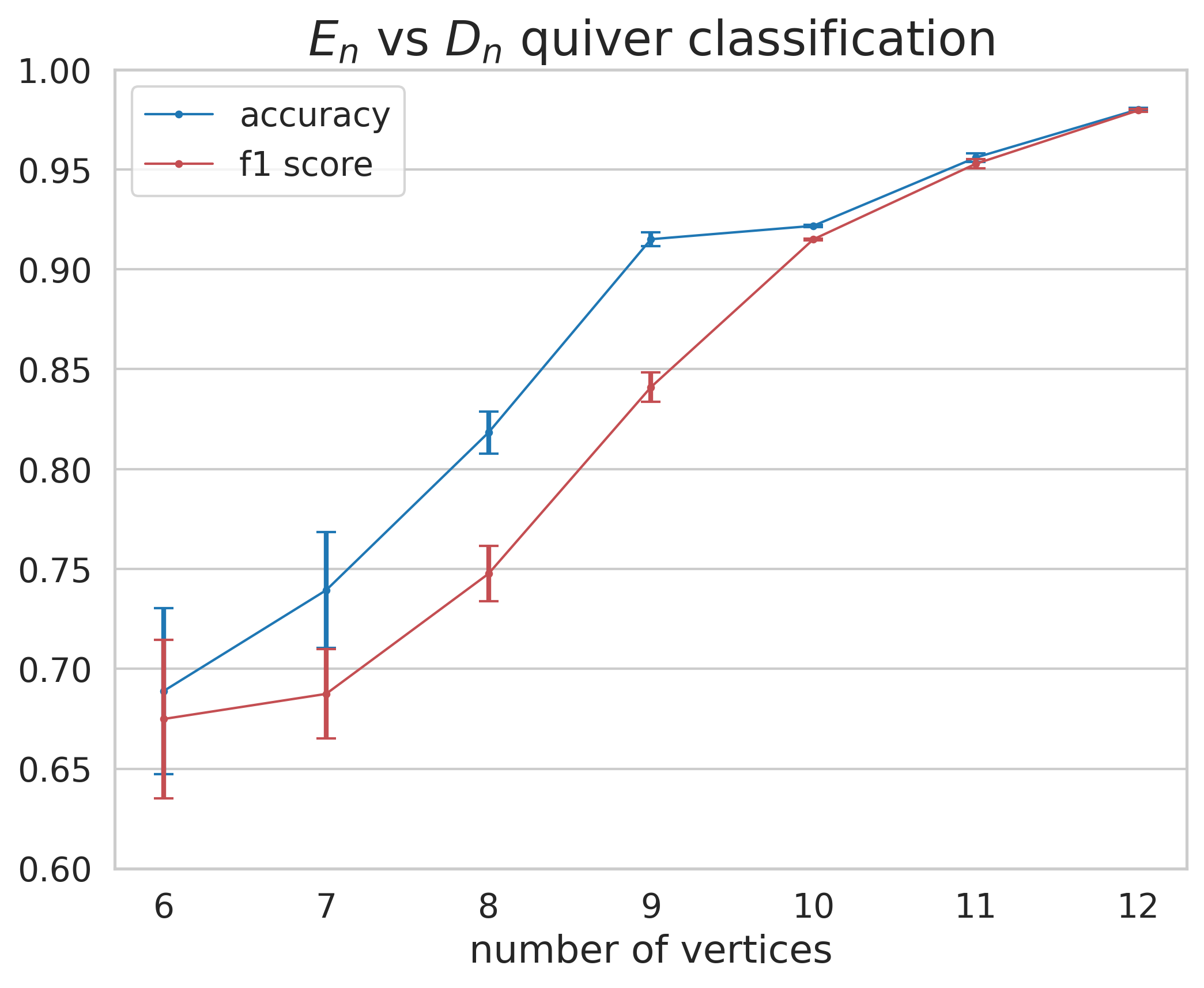}
    \caption{\textbf{(Left)} Performance on the Lattice Path Dataset as a function of the lattice path endpoint (larger endpoint means longer and more paths). As $n$ grows in $n \times n-1$, the training set size increases but the problem may also grow harder. \textbf{(Center)} Performance on the type $E$ versus type $D$ quiver classification task as a function of the depth, which must be specified for type $E$ quivers on $n = 10, 11, 12$ vertices, and \textbf{(Right)} the number of vertices $n$. } 
    \label{fig:dependence_on_n}
    \vspace{-.3cm}
\end{figure}

\section{Dependence on $n$} 

Many problems in algebraic combinatorics have a natural dependence on a parameter $n$ (e.g., permutations are parameterized by the number of elements that they permute). We have chosen to structure datasets in the ACD Repo to reflect this, with the majority of datasets taking the form of a series of datasets $\{D_n\}_{n \geq 1}$. We provide a few values of $n$ and, in many cases, the code to generate others. Generally, there are two properties that change as $n \rightarrow \infty$. First, the size of $D_n$ grows as $n$ grows. The rate of growth depends on the specific problem, with many $|D_n|$ growing exponentially (such as those datasets that depend on the number of permutations of $n$). On the other hand, the problems can also become harder in various ways as $n$ grows. 

Experimentally we have found that larger values of $n$ tend to lead to better model performance regardless of this potential increase in difficulty. For example, we ran 5 2-layer MLP models for $500$ epochs on the lattice path datasets corresponding to grids of size $6\times 5, \dots, 13 \times 12$. We see in Figure \ref{fig:dependence_on_n} (left) that with an interesting exception of moving from $7 \times 6$ to $8 \times 7$, performance across a range of dimensions improves as $n$ grows. There are exceptions, however. We looked at sampling from greater depth when exploring the quiver mutation equivalence dataset (\cref{subsect-quivers}). This effectively means allowing a greater number of mutations to be applied to the initial quiver when generating the dataset. As shown in Figure \ref{fig:dependence_on_n} (center), we find that performance somewhat degrades even though the size of the datasets increases. We suspect that exploration of the complexity of these problems (where it is known) might be an avenue for shedding light on this phenomenon. 

\section{Dataset details}
\label{appendix:dataset_details}

All datasets are stored as \texttt{.txt} files with one datum instance per line. In this section we will describe each file and explain how to interpret it. Functions capable of loading and parsing each file are available on the GitHub page. 


\subsection{Characters of Irreducible Representations of the Symmetric Group}

Since the conjugacy classes of the symmetric group $S_n$ are indexed by integer partitions of $n$, characters are constant on conjugacy classes, and the irreducible representations of $S_n$ are also indexed by integer partitions of $n$, the task is to use a pair of integer partitions of $n$ to predict the character of the corresponding irreducible representation of the symmetric group.

Within each file, two integer partitions are provided followed by an integer corresponding to the character. For instance, the line

\begin{verbatim}
[3,1,1],[2,2,1],-2
\end{verbatim}

says that the character $\chi^{3,1,1}_{2,2,1} = -2$. 

In all cases the characters are heavily concentrated around 0 with very long tails. This likely contributes to the difficulty of the task and could be overcome with some simple pre- and post-processing. We have not chosen to do this in our baselines.

\textbf{Characters of $S_{18}$}

There are $118,580$ training examples and $29,645$ test examples. The maximum character value is $16,336,320$. The minimum character value is $-1,223,040$.

\begin{figure}[h!]
    \centering
    \includegraphics[scale=0.35]{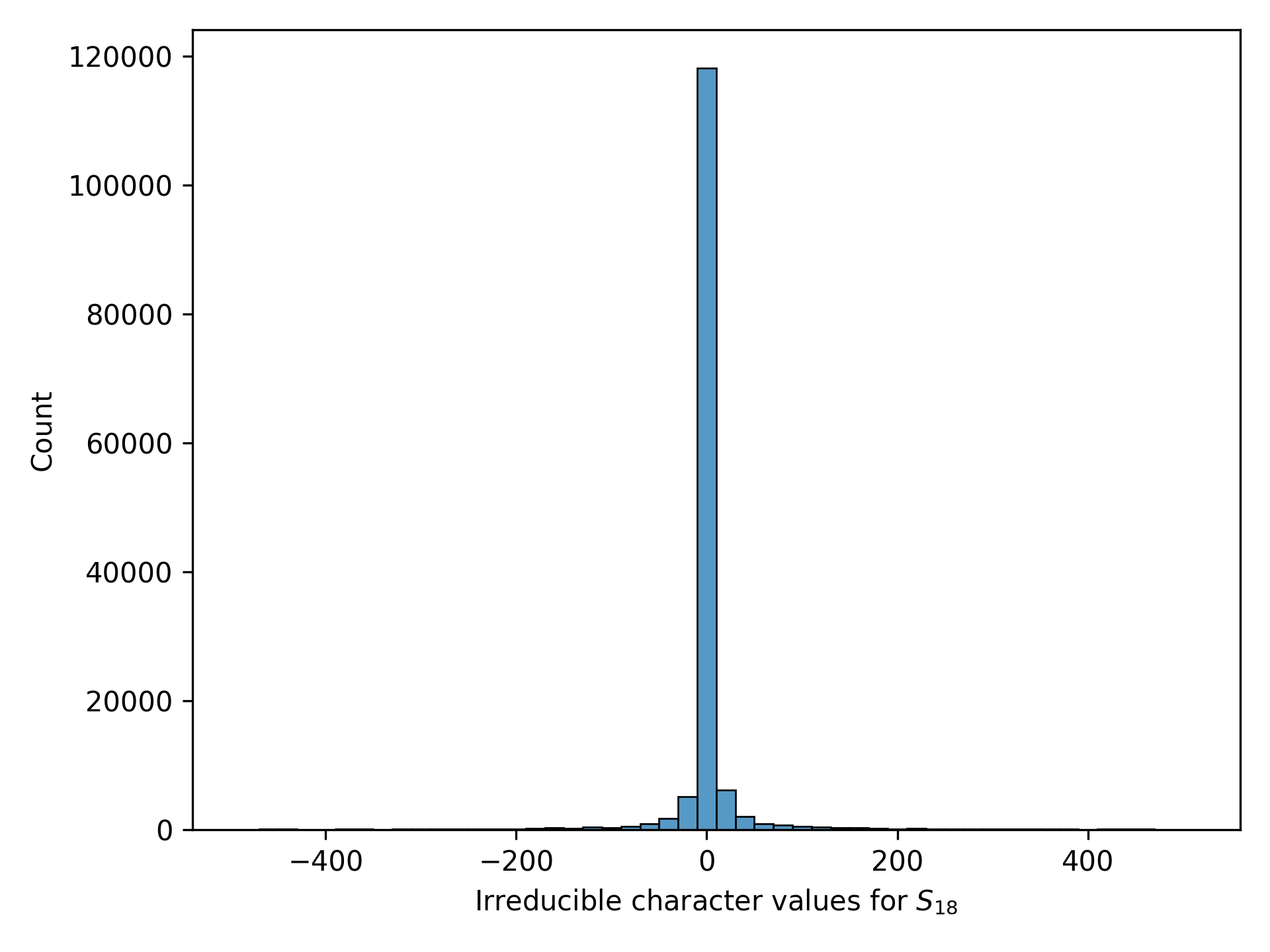}
    \caption{Histogram of $S_{18}$ characters within the interval $[-500,500]$}
    \label{fig:symmetric_histogram_18}
\end{figure}

\textbf{Characters of $S_{20}$}

There are $298,661$ training examples and $74,819$ test examples. The maximum character value is $249,420,600$. The minimum character value is $-17,592,960$.

\begin{figure}[h!]
    \centering
    \includegraphics[scale=0.35]{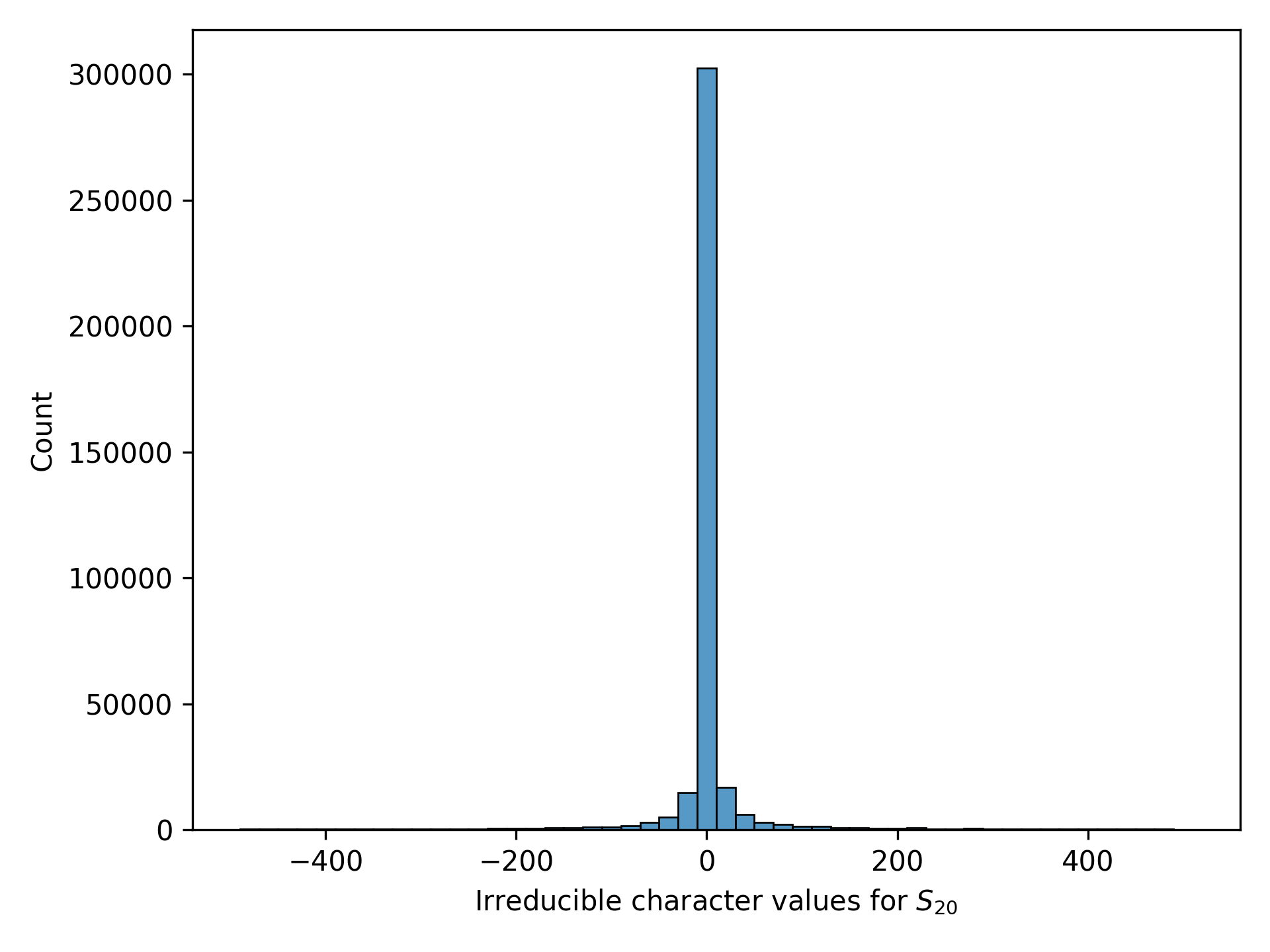}
    \caption{Histogram of $S_{20}$ characters within the interval $[-500,500]$}
    \label{fig:symmetric_histogram_20}
\end{figure}

\textbf{Characters of $S_{22}$}

There are $763,109$ training examples and $190,726$ test examples. The maximum character value is $5,462,865,408$. The minimum character value is $-279,734,796$.

\begin{figure}[h!]
    \centering
    \includegraphics[scale=0.35]{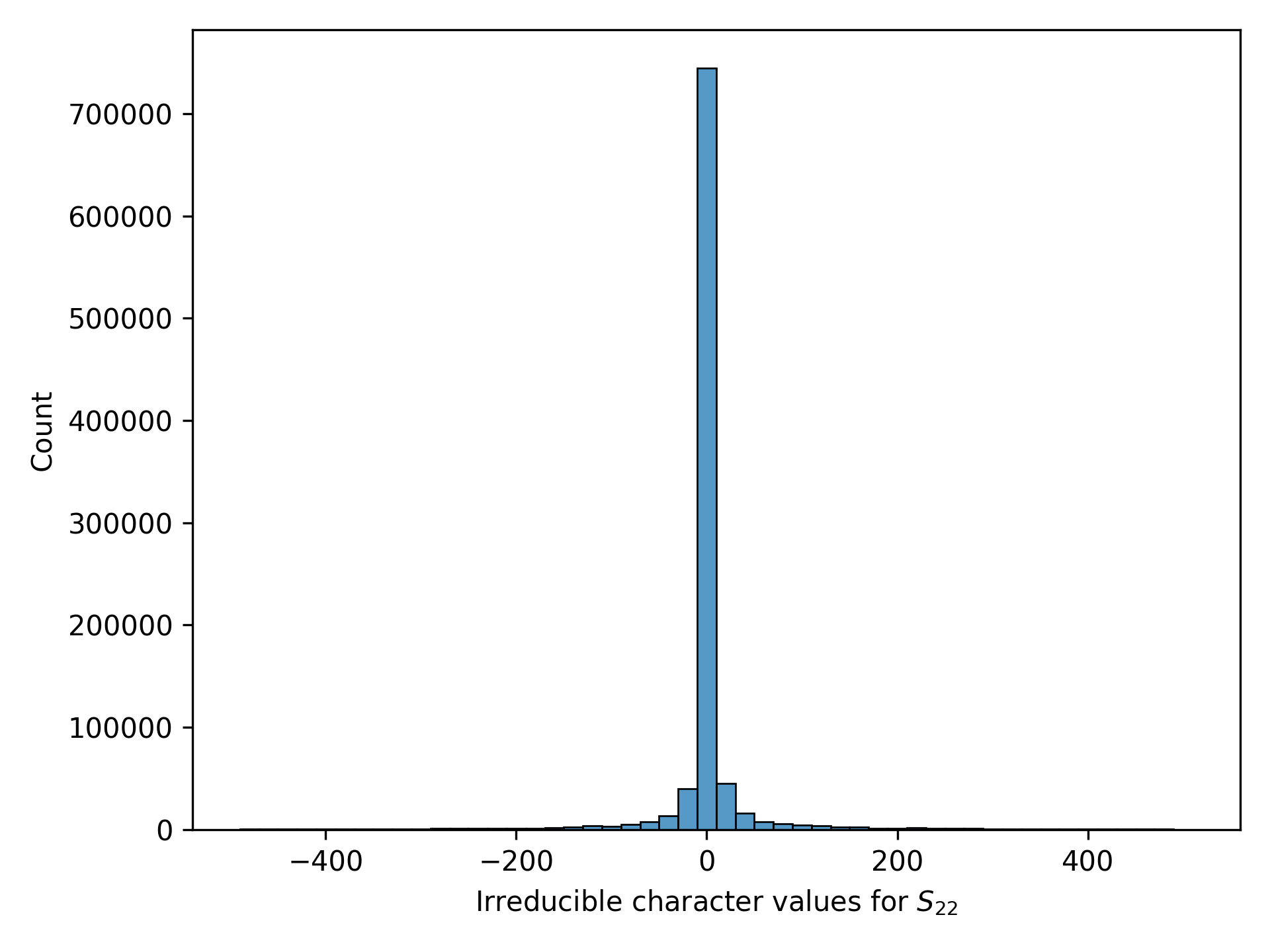}
    \caption{Histogram of $S_{22}$ characters within the interval $[-500,500]$}
    \label{fig:symmetric_histogram_22}
\end{figure}

 Sage \cite{sage} was used to calculate character values. Data generation scripts can be found on the GitHub page.


\subsection{The mHeight Function}

\begin{figure}[h!]
    \centering
    \includegraphics[scale=0.15]{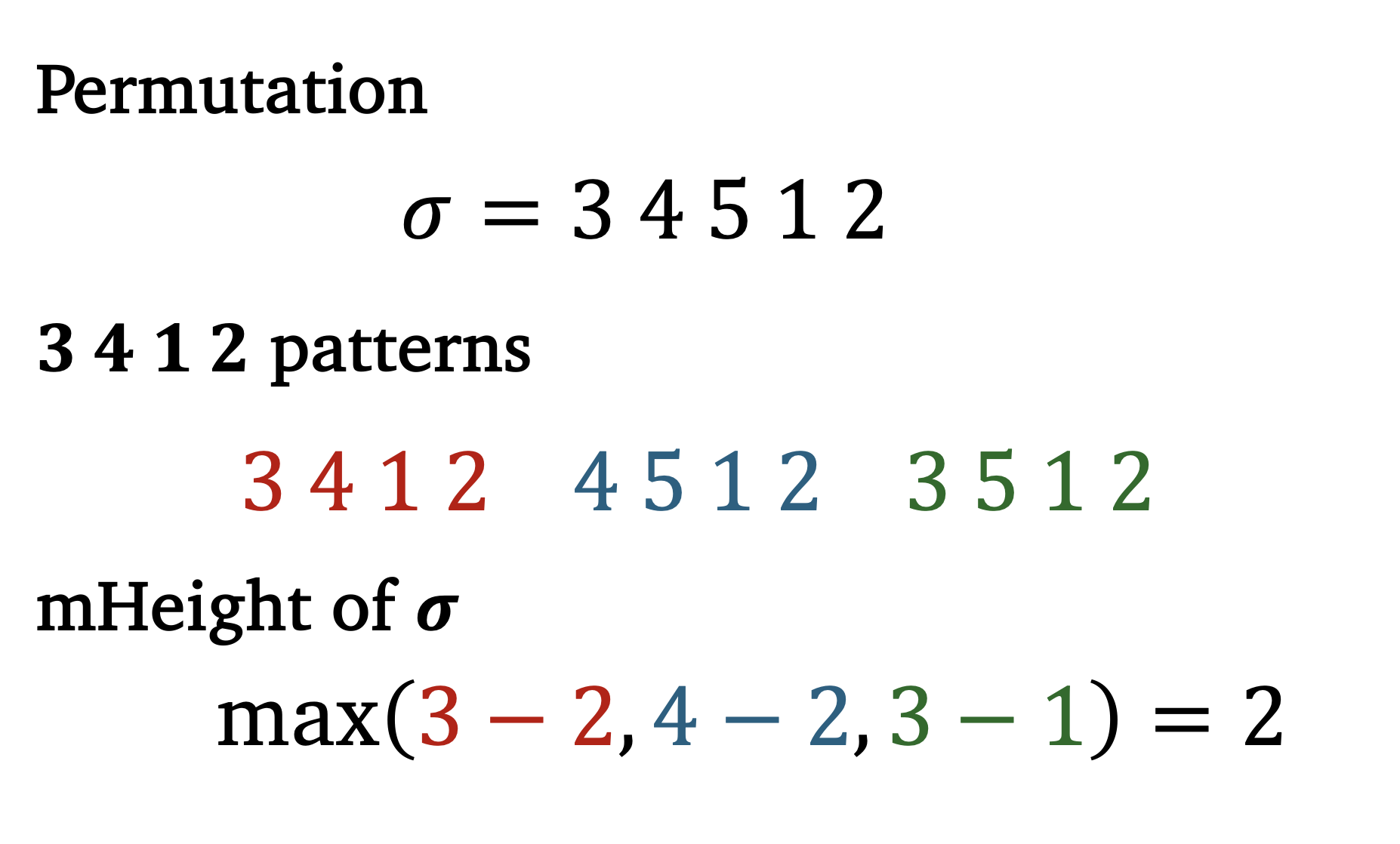}
    \caption{An example of the calculation of mHeight on a permutation.}
    \label{fig:mHeight-example}
\end{figure}

This dataset contains permutations labeled by their mHeight. An example calculation of mHeight is shown in \cref{fig:mHeight-example}. Permutations are written in 1-line notation followed by the symbol `;' and then the value of the mHeight function on the permutation. So 
\begin{verbatim}
    (6, 8, 7, 5, 4, 9, 3, 0, 1, 2);1
\end{verbatim}
can be read as saying that the permutation $6\;8\;7\;5\;4\;9\;3\;0\;1\;2$ has mHeight 1. 

We provide datasets for permutations of size $n = 8,9,10$. The files are called:
\begin{itemize}
\item \begin{verbatim}mHeight_8_train.txt \end{verbatim}
\item \begin{verbatim}mHeight_8_test.txt \end{verbatim}
\item \begin{verbatim}mHeight_9_train.txt \end{verbatim}
\item \begin{verbatim}mHeight_9_test.txt \end{verbatim}
\item \begin{verbatim}mHeight_10_train.txt \end{verbatim}
\item \begin{verbatim}mHeight_10_test.txt \end{verbatim}
\end{itemize}

The dataset was generated using a Python script which can be found on the GitHub page. Dataset statistics can be found in \cref{tab:mheight-stats-8}-\cref{tab:mheight-stats-10}.

\begin{table*}\centering
\begin{tabular}{@{}lccccc@{}}\toprule
& \footnotesize{0} & \footnotesize{1}  & \footnotesize{2} & \footnotesize{3} & \footnotesize{4} \\  
\midrule
\hspace{10pt}\scriptsize{Training set} & \scriptsize{$6,716$} & \scriptsize{$508$} &  \scriptsize{$78$} &  \scriptsize{$9$} &  \scriptsize{$1$} \\
\hspace{10pt}\scriptsize{Testing set} & \scriptsize{$1,672$} & \scriptsize{$136$} &  \scriptsize{$18$} &  \scriptsize{$3$} &  \scriptsize{$0$}\\
\bottomrule
\end{tabular}
\caption{\label{tab:mheight-stats-8}Statistics for the mHeight dataset for $n = 8$.}
\vspace{-.3cm}
\end{table*}

\begin{table*}\centering
\begin{tabular}{@{}lcccccc@{}}\toprule
& \footnotesize{0} & \footnotesize{1}  & \footnotesize{2} & \footnotesize{3} & \footnotesize{4} & \footnotesize{5} \\  
\midrule
\hspace{10pt}\scriptsize{Training set} & \scriptsize{$49,092$} & \scriptsize{$3,161$} &  \scriptsize{$524$} &  \scriptsize{$77$} &  \scriptsize{$9$} & \scriptsize{$1$} \\
\hspace{10pt}\scriptsize{Testing set} & \scriptsize{$12,317$} & \scriptsize{$759$} &  \scriptsize{$118$} &  \scriptsize{$19$} &  \scriptsize{$3$}  &  \scriptsize{$0$}\\
\bottomrule
\end{tabular}
\caption{\label{tab:mheight-stats-9}Statistics for the mHeight dataset for $n = 9$.}
\vspace{-.3cm}
\end{table*}

\begin{table*}\centering
\begin{tabular}{@{}lccccccc@{}}\toprule
& \footnotesize{0} & \footnotesize{1}  & \footnotesize{2} & \footnotesize{3} & \footnotesize{4} & \footnotesize{5} & 6 \\  
\midrule
\hspace{10pt}\scriptsize{Training set} & \scriptsize{$352,494$} & \scriptsize{$17,952$} &  \scriptsize{$3,079$} &  \scriptsize{$502$} &  \scriptsize{$74$} & \scriptsize{$10$} & \scriptsize{$1$}\\
\hspace{10pt}\scriptsize{Testing set} & \scriptsize{$88,058$} & \scriptsize{$4,503$} &  \scriptsize{$803$} &  \scriptsize{$140$} &  \scriptsize{$22$} & \scriptsize{$2$} & \scriptsize{$0$}\\
\bottomrule
\end{tabular}
\caption{\label{tab:mheight-stats-10}Statistics for the mHeight dataset for $n = 10$.}
\vspace{-.3cm}
\end{table*}

\subsection{Semistandard Young Tableaux for the Grassmannian} 

This dataset relates to a cluster algebra associated with the Grassmann manifold $\text{Gr}(k,n)$. Each cluster variable is indexed by a rectangular SSYT with $k$ rows with entries drawn from $\{1,\dots,n\}$. The {\emph{rank}} of these rectangular SSYT (and the rank of the associated cluster variable) is given by their number of columns. Following \cite{cheung2022clustering}, in this dataset we focus on $\text{Gr}(3,12)$ and hence look at rectangular SSYT with 3 rows filled with entries drawn from $\{1,\dots,12\}$. We further restrict to rank 4 SSYT. This leaves us with a collection of $3 \times 4$ arrays whose entries increase weakly across rows and strictly down columns. 

\begin{figure}[h!]
    \centering
    \includegraphics[scale=0.15]{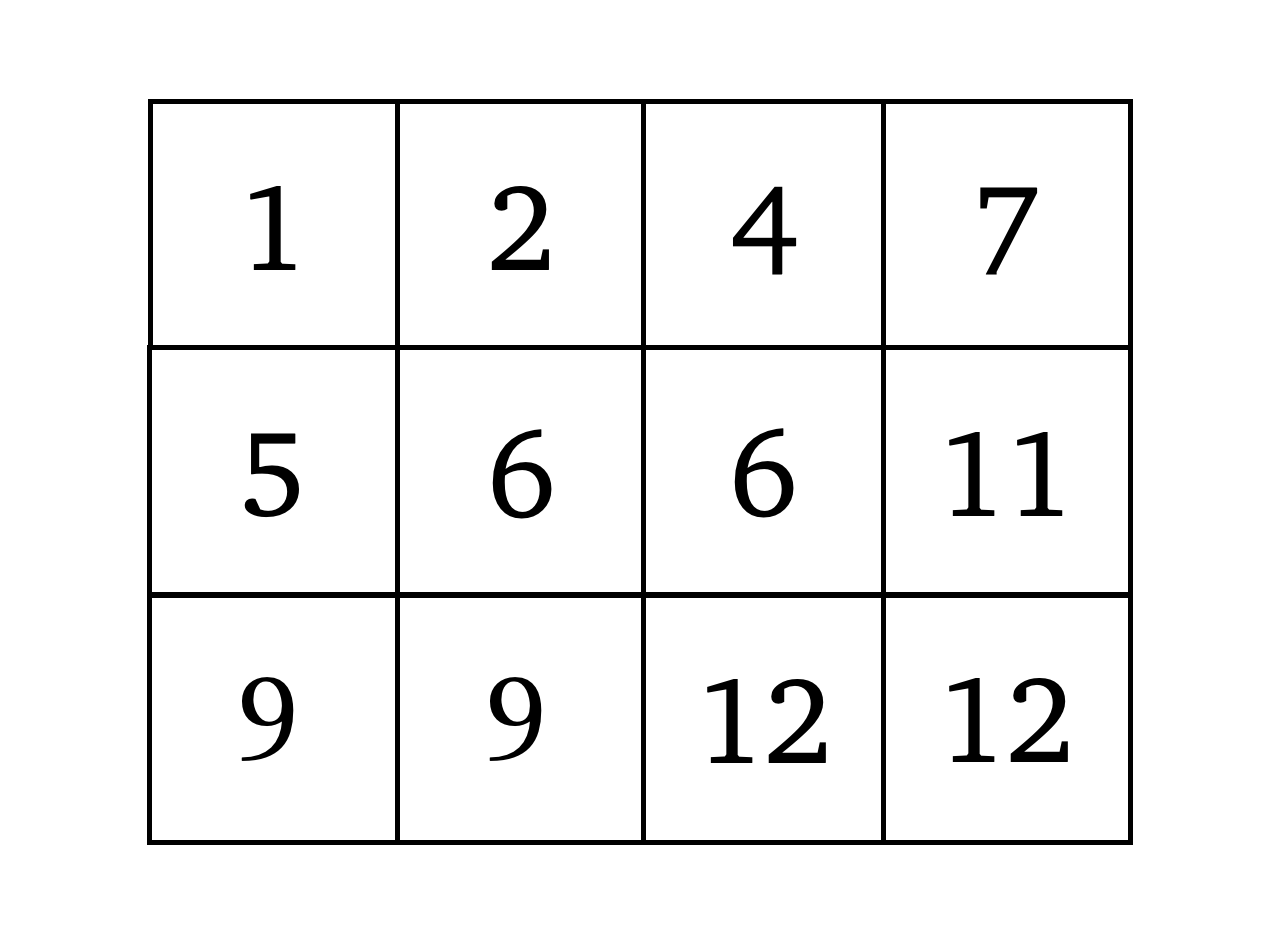}
    \includegraphics[scale=0.15]{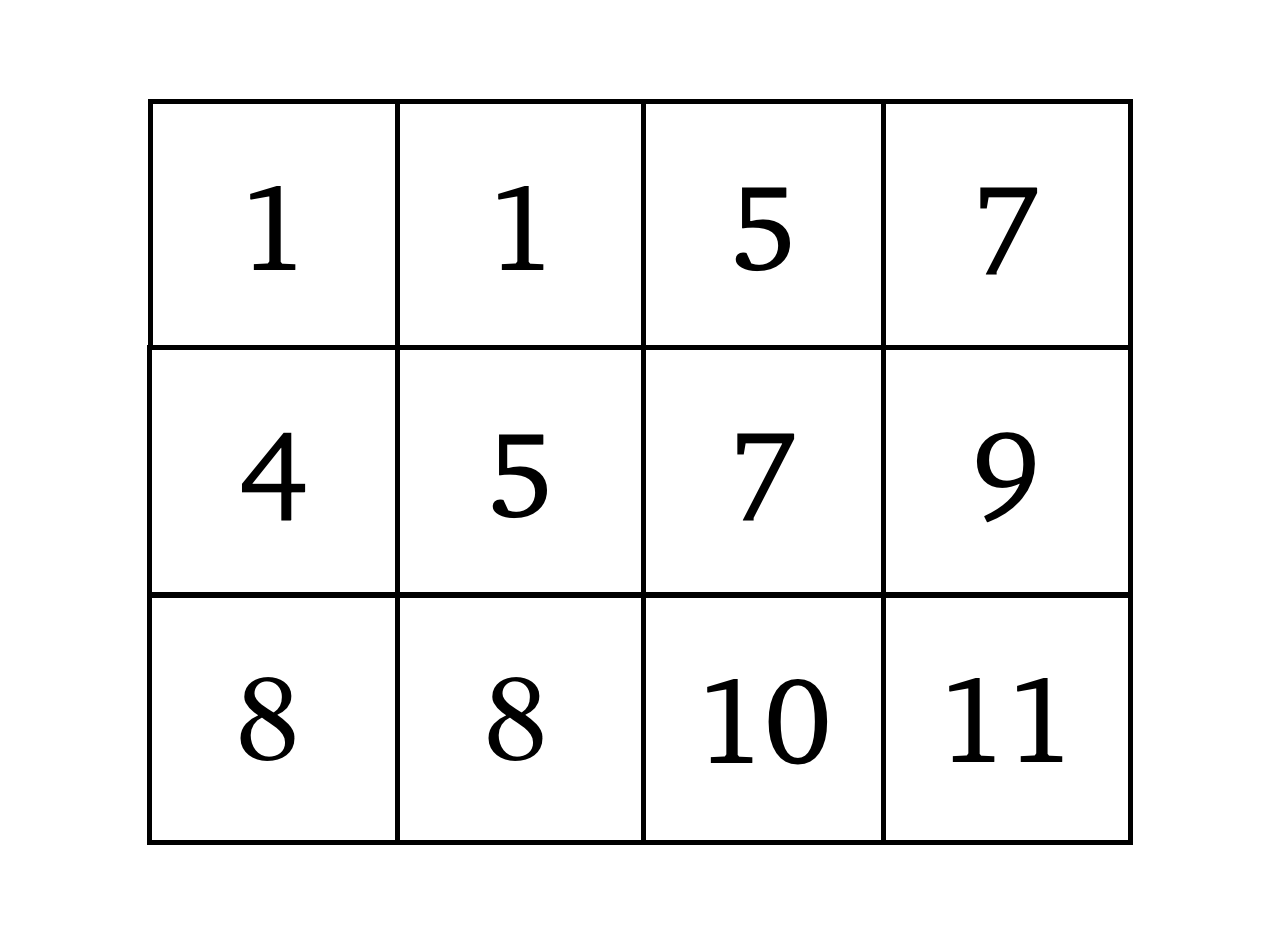}
    \caption{An example of a valid \textbf{(left)} and invalid \textbf{(right)} tableau from the Grassmannian cluster algebra dataset.}
    \label{fig:grassmannian-cluster-valid-invalid}
\end{figure}

To give two examples, the SSYT in \cref{fig:grassmannian-cluster-valid-invalid} are valid (left) and invalid (right). Note that both are genuine SSYT of shape $3 \times 4$ with entries from $\{1, \dots, 12\}$.

The dataset consists of a collection of rectangular SSYT each with a label indicating whether it indexes a cluster variable or not. Those that do not index a cluster variable are labeled with a `0' and those that do are labeled with a `1'. The valid examples are drawn from \cite{cheung2022clustering} and can be obtained from \url{https://github.com/edhirst/GrassmanniansML/}.  We generated our own negative examples because we found that the model learned some spurious correlations as a result of the sampling strategy used in \cite{cheung2022clustering}. To sample random rectangular SSYT, we took advantage of the {\texttt{random\_element\(\)}} method in the `Tableaux' class in Sage.

The datasets, which can be found on the GitHub page, are contained in files named: 
\begin{itemize}
\item \begin{verbatim}3_4_12_invalid_train.txt\end{verbatim}
\item \begin{verbatim}3_4_12_invalid_test.txt\end{verbatim}
\item \begin{verbatim}3_4_12_valid_test.txt\end{verbatim}
\item \begin{verbatim}3_4_12_valid_train.txt\end{verbatim}
\end{itemize}

In the files we use braces $[$ and $]$ to demarcate rows of the diagram, so that

\begin{verbatim}[[1, 2, 4, 7], [5, 6, 6, 11], [9, 9, 12, 12]]\end{verbatim}

corresponds to the tableau in Figure~\ref{fig:grassmannian-cluster-valid-invalid}, left.
Dataset statistics can be found in \cref{tab:grassmannian-stats}.

\begin{table*}\centering
\begin{tabular}{@{}lcccccccc@{}}\toprule
& \footnotesize{Cluster variable} & \footnotesize{Not cluster variable}  & \footnotesize{Total} \\  
\midrule

\hspace{10pt}\scriptsize{Training set} & \scriptsize{$74,329$} & \scriptsize{$74,329$} &  \scriptsize{$148,658$} \\
\hspace{10pt}\scriptsize{Testing set} & \scriptsize{$18,582$} & \scriptsize{$18,582$} &  \scriptsize{$37,164$}\\
\bottomrule
\end{tabular}
\caption{\label{tab:grassmannian-stats}Statistics of the Grassmannian cluster algebra dataset.}
\vspace{-.3cm}
\end{table*}


\subsection{Kazhdan-Lusztig Polynomials}

Kazhdan-Lusztig polynomials are integer polynomials in a variable $q$ which are parameterized by two permutations. The tasks associated with this dataset are to predict the different coefficients of the polynomial (organized by monomial degree). Thus the input is two permutations $v,w$ and the output is a sequence of integers which are the coefficients of the polynomial. For instance, if $v = 0\;2\;1\;3\;5\;4\;6\;9\;7\;8$ and $w=2\;3\;0\;5\;9\;6\;7\;8\;1\;4$ then
\begin{equation*}
P_{v,w}(q) = 4q^4+12q^3+13q^2+6q+1.
\end{equation*}
and this is written as the line
\begin{verbatim}
    0213546978 2305967814 1,6,13,12,4. 
\end{verbatim}
Datasets were generated using C code from Greg Warrington's website \cite{warringtonwebsite}, computing $P_{v,w}(q)$ for all pairs of permutations of size $n = 5, 6, 7$.

The files we provide are: 
\begin{itemize}
\item \begin{verbatim}kl-polynomials_5_train.txt \end{verbatim}
\item \begin{verbatim}kl-polynomials_5_test.txt \end{verbatim}
\item \begin{verbatim}kl-polynomials_6_train.txt \end{verbatim}
\item \begin{verbatim}kl-polynomials_6_test.txt \end{verbatim}
\item \begin{verbatim}kl-polynomials_7_train.txt \end{verbatim}
\item \begin{verbatim}kl-polynomials_7_test.txt \end{verbatim}
\end{itemize}

Statistics can be found in \cref{tab:kl-n5-0}-\cref{tab:kl-n7-4}.

\begin{table*}\centering
\begin{tabular}{@{}lcc@{}}\toprule
\footnotesize{Coefficient} & \footnotesize{Train} & \footnotesize{Test}   \\  
\midrule
\hspace{10pt}\scriptsize{0} & \scriptsize{$8,496$} & \scriptsize{$2,123$}    \\
\hspace{10pt}\scriptsize{1} & \scriptsize{$3,024$} & \scriptsize{$757$}  \\
\bottomrule
\end{tabular}
\caption{\label{tab:kl-n5-0}Statistics for the constant term in the $n=5$ KL polynomial dataset.}
\vspace{-.3cm}
\end{table*}

\begin{table*}\centering
\begin{tabular}{@{}lcc@{}}\toprule
\footnotesize{Coefficient} & \footnotesize{Train} & \footnotesize{Test}   \\  
\midrule
\hspace{10pt}\scriptsize{0} & \scriptsize{$11,219$} & \scriptsize{$2,793$}  \\
\hspace{10pt}\scriptsize{1} & \scriptsize{$267$} & \scriptsize{$77$}\\
\hspace{10pt}\scriptsize{2} & \scriptsize{$34$} & \scriptsize{$10$} \\
\bottomrule
\end{tabular}
\caption{\label{tab:kl-n5-1}Statistics for the coefficient on $q$ in the $n=5$ KL polynomial dataset.}
\vspace{-.3cm}
\end{table*}

\begin{table*}\centering
\begin{tabular}{@{}lcc@{}}\toprule
\footnotesize{Coefficient} & \footnotesize{Train} & \footnotesize{Test}   \\  
\midrule
\hspace{10pt}\scriptsize{0} & \scriptsize{$11,514$} & \scriptsize{$2,876$} \\
\hspace{10pt}\scriptsize{1} & \scriptsize{$6$} & \scriptsize{$4$} \\
\bottomrule
\end{tabular}
\caption{\label{tab:kl-n5-2}Statistics for the coefficient on $q^2$ in the $n=5$ KL polynomial dataset.}
\vspace{-.3cm}
\end{table*}

\begin{table*}\centering
\begin{tabular}{@{}lcc@{}}\toprule
\footnotesize{Coefficient} & \footnotesize{Train} & \footnotesize{Test}   \\  
\midrule
\hspace{10pt}\scriptsize{0} & \scriptsize{$336,071$} & \scriptsize{$83,922$} \\
\hspace{10pt}\scriptsize{1} & \scriptsize{$78,649$} & \scriptsize{$19,758$} \\
\bottomrule
\end{tabular}
\caption{\label{tab:kl-n6-0}Statistics for the constant term in the $n=6$ KL polynomial dataset.}
\vspace{-.3cm}
\end{table*}

\begin{table*}\centering
\begin{tabular}{@{}lcc@{}}\toprule
\footnotesize{Coefficient} & \footnotesize{Train} & \footnotesize{Test}   \\  
\midrule
\hspace{10pt}\scriptsize{0} & \scriptsize{$397,386$} & \scriptsize{$99,354$} \\
\hspace{10pt}\scriptsize{1} & \scriptsize{$13,253$} & \scriptsize{$3,311$} \\
\hspace{10pt}\scriptsize{2} & \scriptsize{$3,483$} & \scriptsize{$883$} \\
\hspace{10pt}\scriptsize{3} & \scriptsize{$535$} & \scriptsize{$117$} \\
\hspace{10pt}\scriptsize{4} & \scriptsize{$63$} & \scriptsize{$15$} \\
\bottomrule
\end{tabular}
\caption{\label{tab:kl-n6-1}Statistics for the coefficient on $q$ in the $n=6$ KL polynomial dataset.}
\vspace{-.3cm}
\end{table*}

\begin{table*}\centering
\begin{tabular}{@{}lcc@{}}\toprule
\footnotesize{Coefficient} & \footnotesize{Train} & \footnotesize{Test}   \\  
\midrule
\hspace{10pt}\scriptsize{0} & \scriptsize{$412,707$} & \scriptsize{$103,177$} \\
\hspace{10pt}\scriptsize{1} & \scriptsize{$1,705$} & \scriptsize{$441$} \\
\hspace{10pt}\scriptsize{2} & \scriptsize{$242$} & \scriptsize{$46$} \\
\hspace{10pt}\scriptsize{3} & \scriptsize{$40$} & \scriptsize{$8$} \\
\hspace{10pt}\scriptsize{4} & \scriptsize{$26$} & \scriptsize{$8$} \\
\bottomrule
\end{tabular}
\caption{\label{tab:kl-n6-2}Statistics for the coefficient on $q^2$ in the $n=6$ KL polynomial dataset.}
\vspace{-.3cm}
\end{table*}

\begin{table*}\centering
\begin{tabular}{@{}lcc@{}}\toprule
\footnotesize{Coefficient} & \footnotesize{Train} & \footnotesize{Test}   \\  
\midrule
\hspace{10pt}\scriptsize{0} & \scriptsize{$414,688$} & \scriptsize{$103,670$} \\
\hspace{10pt}\scriptsize{1} & \scriptsize{$32$} & \scriptsize{$10$} \\
\bottomrule
\end{tabular}
\caption{\label{tab:kl-n6-3}Statistics for the coefficient on $q^3$ in the $n=6$ KL polynomial dataset.}
\vspace{-.3cm}
\end{table*}

\begin{table*}\centering
\begin{tabular}{@{}lcc@{}}\toprule
\footnotesize{Coefficient} & \footnotesize{Train} & \footnotesize{Test}   \\  
\midrule
\hspace{10pt}\scriptsize{0} & \scriptsize{$17,479,910$} & \scriptsize{$4,370,771$} \\
\hspace{10pt}\scriptsize{1} & \scriptsize{$2,841,370$} & \scriptsize{$709,549$} \\
\bottomrule
\end{tabular}
\caption{\label{tab:kl-n7-0}Statistics for the constant term in the $n=7$ KL polynomial dataset.}
\vspace{-.3cm}
\end{table*}

\begin{table*}\centering
\begin{tabular}{@{}lcc@{}}\toprule
\footnotesize{Coefficient} & \footnotesize{Train} & \footnotesize{Test}   \\  
\midrule
\hspace{10pt}\scriptsize{0} & \scriptsize{$19,291,150$} & \scriptsize{$4,822,214$} \\
\hspace{10pt}\scriptsize{1} & \scriptsize{$660,600$} & \scriptsize{$165,768$} \\
\hspace{10pt}\scriptsize{2} & \scriptsize{$266,591$} & \scriptsize{$66,593$} \\
\hspace{10pt}\scriptsize{3} & \scriptsize{$80,173$} & \scriptsize{$19,963$} \\
\hspace{10pt}\scriptsize{4} & \scriptsize{$18,834$} & \scriptsize{$4,762$} \\
\hspace{10pt}\scriptsize{5} & \scriptsize{$3,221$} & \scriptsize{$819$} \\
\hspace{10pt}\scriptsize{6} & \scriptsize{$711$} & \scriptsize{$201$} \\
\bottomrule
\end{tabular}
\caption{\label{tab:kl-n7-1}Statistics for the coefficient on $q$ in the $n=7$ KL polynomial dataset.}
\vspace{-.3cm}
\end{table*}

\begin{table*}\centering
\begin{tabular}{@{}lcc@{}}\toprule
\footnotesize{Coefficient} & \footnotesize{Train} & \footnotesize{Test}   \\  
\midrule
\hspace{10pt}\scriptsize{0} & \scriptsize{$20,072,738$} & \scriptsize{$5,017,962$} \\
\hspace{10pt}\scriptsize{1} & \scriptsize{$170,412$} & \scriptsize{$42,748$} \\
\hspace{10pt}\scriptsize{2} & \scriptsize{$46,226$} & \scriptsize{$11,568$} \\
\hspace{10pt}\scriptsize{3} & \scriptsize{$16,227$} & \scriptsize{$4,021$} \\
\hspace{10pt}\scriptsize{4} & \scriptsize{$7,621$} & \scriptsize{$1,905$} \\
\hspace{10pt}\scriptsize{5} & \scriptsize{$4,023$} & \scriptsize{$1,065$} \\
\hspace{10pt}\scriptsize{6} & \scriptsize{$1,287$} & \scriptsize{$349$} \\
\hspace{10pt}\scriptsize{7} & \scriptsize{$1,153$} & \scriptsize{$287$} \\
\hspace{10pt}\scriptsize{8} & \scriptsize{$785$} & \scriptsize{$183$} \\
\hspace{10pt}\scriptsize{9} & \scriptsize{$350$} & \scriptsize{$86$} \\
\hspace{10pt}\scriptsize{10} & \scriptsize{$152$} & \scriptsize{$40$} \\
\hspace{10pt}\scriptsize{11} & \scriptsize{$139$} & \scriptsize{$37$} \\
\hspace{10pt}\scriptsize{12} & \scriptsize{$121$} & \scriptsize{$47$} \\
\hspace{10pt}\scriptsize{13} & \scriptsize{$42$} & \scriptsize{$22$} \\
\hspace{10pt}\scriptsize{14} & \scriptsize{$4$} & \scriptsize{$1$} \\
\bottomrule
\end{tabular}
\caption{\label{tab:kl-n7-2}Statistics for the coefficient on $q^2$ in the $n=7$ KL polynomial dataset.}
\vspace{-.3cm}
\end{table*}

\begin{table*}\centering
\begin{tabular}{@{}lcc@{}}\toprule
\footnotesize{Coefficient} & \footnotesize{Train} & \footnotesize{Test}   \\  
\midrule
\hspace{10pt}\scriptsize{0} & \scriptsize{$20,291,535$} & \scriptsize{$507,2831$} \\
\hspace{10pt}\scriptsize{1} & \scriptsize{$22,094$} & \scriptsize{$5,498$} \\
\hspace{10pt}\scriptsize{2} & \scriptsize{$4,779$} & \scriptsize{$1,213$} \\
\hspace{10pt}\scriptsize{3} & \scriptsize{$1,660$} & \scriptsize{$442$} \\
\hspace{10pt}\scriptsize{4} & \scriptsize{$590$} & \scriptsize{$146$} \\
\hspace{10pt}\scriptsize{5} & \scriptsize{$195$} & \scriptsize{$61$} \\
\hspace{10pt}\scriptsize{6} & \scriptsize{$206$} & \scriptsize{$50$} \\
\hspace{10pt}\scriptsize{7} & \scriptsize{$115$} & \scriptsize{$37$} \\
\hspace{10pt}\scriptsize{8} & \scriptsize{$34$} & \scriptsize{$14$} \\
\hspace{10pt}\scriptsize{9} & \scriptsize{$26$} & \scriptsize{$6$} \\
\hspace{10pt}\scriptsize{10} & \scriptsize{$24$} & \scriptsize{$8$} \\
\hspace{10pt}\scriptsize{11} & \scriptsize{$18$} & \scriptsize{$14$} \\
\hspace{10pt}\scriptsize{15} & \scriptsize{$4$} & \scriptsize{$1$} \\
\bottomrule
\end{tabular}
\caption{\label{tab:kl-n7-3}Statistics for the coefficient on $q^3$ in the $n=7$ KL polynomial dataset.}
\vspace{-.3cm}
\end{table*}

\begin{table*}\centering
\begin{tabular}{@{}lcc@{}}\toprule
\footnotesize{Coefficient} & \footnotesize{Train} & \footnotesize{Test}   \\  
\midrule
\hspace{10pt}\scriptsize{0} & \scriptsize{$17,479,910$} & \scriptsize{$4,370,771$} \\
\hspace{10pt}\scriptsize{1} & \scriptsize{$2,841,370$} & \scriptsize{$709,549$} \\
\bottomrule
\end{tabular}
\caption{\label{tab:kl-n7-4}Statistics for the coefficient on $q^4$ in the $n=7$ KL polynomial dataset.}
\vspace{-.3cm}
\end{table*}


\subsection{The Robinson–Schensted-Knuth Correspondence}


The Robinson–Schensted-Knuth correspondence is a bijection between permutations and pairs of standard Young tableaux. For our initial baselines we use Young tableau pairs as input and the corresponding permutation as output, but the dataset could be used in the opposite direction.

\begin{figure}[h!]
    \centering
    \includegraphics[scale=0.15]{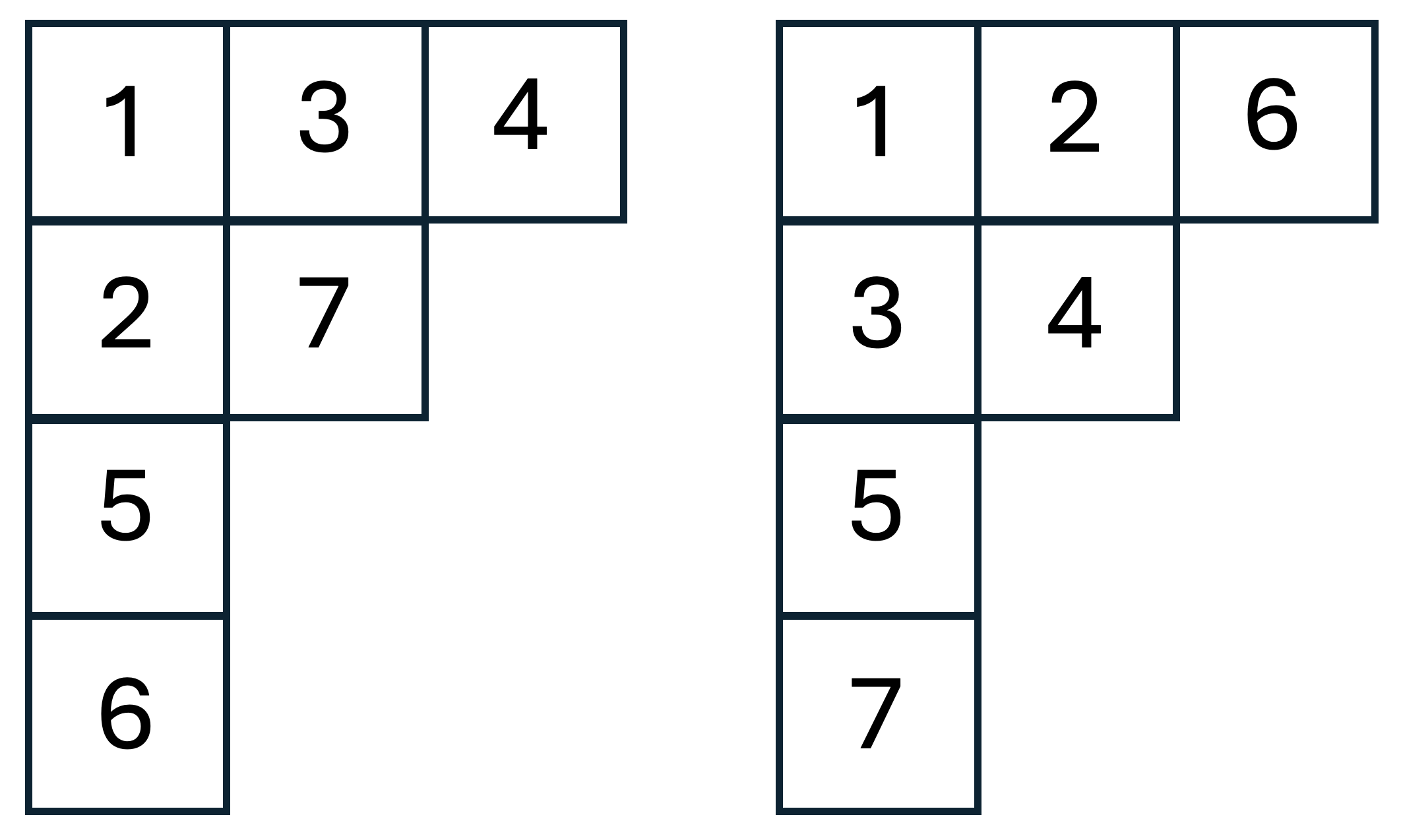}
    \caption{The Young tableau pair that are obtained by applying the Robinson-Schensted-Knuth algorithm to the permutation $6\;7\;2\;5\;3\;4\;1$.}
    \label{fig:rsk-output}
\end{figure}

The permutations are stored in files starting with {\texttt{input\_permutation}} and the pair of tableaux are stored in files labeled by {\texttt{output\_tableau}}. Permutations are stored using their inversion vectors (a binary sequence). Tableau rows are separated by `$[$' and `$]$'. For instance,
\begin{verbatim}
    [[[1, 3, 4], [2, 7], [5], [6]], [[1, 2, 6], [3, 4], [5], [7]]]
\end{verbatim}
corresponds to the pair of Young tableau in Figure \ref{fig:rsk-output}. 

We store files for $n = 8,9,10$. These are named:
\begin{itemize}
\item \begin{verbatim}input_permutations_8_train.csv \end{verbatim}
\item \begin{verbatim}input_permutations_8_test.csv \end{verbatim}
\item \begin{verbatim}input_permutations_9_train.csv \end{verbatim}
\item \begin{verbatim}input_permutations_9_test.csv \end{verbatim}
\item \begin{verbatim}input_permutations_10_train.csv \end{verbatim}
\item \begin{verbatim}input_permutations_10_test.csv \end{verbatim}
\item \begin{verbatim}output_tableau_8_train.csv \end{verbatim}
\item \begin{verbatim}output_tableau_8_test.csv \end{verbatim}
\item \begin{verbatim}output_tableau_9_train.csv \end{verbatim}
\item \begin{verbatim}output_tableau_9_test.csv \end{verbatim}
\item \begin{verbatim}output_tableau_10_train.csv \end{verbatim}
\item \begin{verbatim}output_tableau_10_test.csv \end{verbatim}
\end{itemize}

Sage \cite{sage} was used to generate the tableau pairs corresponding to each permutation. The script is available on the GitHub page. 


\subsection{Schubert Polynomial Structure Constants}

This task involves predicting the structure constants of Schubert polynomials. These polynomials are each indexed by permutations so structure constants are indexed by a triple of permutations. Hence, the input to the model is a triple of permutations $v,w,u$ and the output is an integer. For instance, since we have the relationship
\begin{equation*}
    \mathcal{S}_{1 2  3  5 4}\mathcal{S}_{1 2 3 5 4} = \mathcal{S}_{1  2  3  6  4  5} + \mathcal{S}_{1  2 4  5  3}
\end{equation*}
one data instance is
\begin{verbatim}
    [1,2,3,5,4],[1,2,3,5,4],[1,2,3,6,4,5];1.
\end{verbatim}

We partition the datasets so that the dataset associated with value $n$ has structure constants for pairs of $\mathcal{S}_v$ with $v \in S_n$. Note that there is some repetition since $S_{n-1}$ is a subset of $S_n$. We store files for $n = 4,5,6$. These are named:
\begin{itemize}
\item \begin{verbatim}schubert_structure_coefficients_triples_4_train.txt \end{verbatim}
\item \begin{verbatim}schubert_structure_coefficients_triples_4_test.txt \end{verbatim}
\item \begin{verbatim}schubert_structure_coefficients_triples_5_train.txt \end{verbatim}
\item \begin{verbatim}schubert_structure_coefficients_triples_5_test.txt 
\end{verbatim}
\item \begin{verbatim}schubert_structure_coefficients_triples_6_train.txt \end{verbatim}
\item \begin{verbatim}schubert_structure_coefficients_triples_6_test.txt. \end{verbatim} 
\end{itemize}

Sage \cite{sage} was used to generate and multiply Schubert polynomials for each pair of permutations $\alpha$ and $\beta$ in $S_n$. The basis expansion of $\mathcal{S}_\alpha \star \mathcal{S}_\beta$ was obtained from this and each nonzero term in this expansion was used as an instance.
For any $n$, most structure constants will be zero. To generate a balanced dataset, we computed $\mathcal{S}_\alpha \star \mathcal{S}_\beta$ for all elements in $S_n \times S_n$ and for each $c_{\alpha,\beta}^\gamma \neq 0$, we applied a random number of transpositions (where the number of transpositions was sampled from a geometric distribution) to $\gamma$ to get $\gamma'$, checked that $c_{\alpha,\beta}^{\gamma'} = 0$ and added this to the dataset. Therefore, the dataset contains all non-zero structure constants but only a fraction of zero structure constants.

Dataset statistics can be found in \cref{tab:schubert-stats-4}-\cref{tab:schubert-stats-6}.

\begin{table*}\centering
\begin{tabular}{@{}lcc@{}}\toprule
& \footnotesize{0} & \footnotesize{1}  \\  
\midrule
\hspace{10pt}\scriptsize{Training set} & \scriptsize{$851$} & \scriptsize{$833$}  \\
\hspace{10pt}\scriptsize{Testing set} & \scriptsize{$201$} & \scriptsize{$220$} \\
\bottomrule
\end{tabular}
\caption{\label{tab:schubert-stats-4}Statistics for the Schubert polynomial structure constants dataset for $n = 4$.}
\vspace{-.3cm}
\end{table*}

\begin{table*}\centering
\begin{tabular}{@{}lccc@{}}\toprule
& \footnotesize{0} & \footnotesize{1}  & \footnotesize{2}  \\  
\midrule
\hspace{10pt}\scriptsize{Training set} & \scriptsize{$42,831$} & \scriptsize{$42,619$} &  \scriptsize{$170$}  \\
\hspace{10pt}\scriptsize{Testing set} & \scriptsize{$10,681$} & \scriptsize{$10,680$} &  \scriptsize{$44$}\\
\bottomrule
\end{tabular}
\caption{\label{tab:schubert-stats-5}Statistics for the Schubert polynomial structure constants dataset for $n = 5$.}
\end{table*}

\begin{table*}\centering
\begin{tabular}{@{}lcccccc@{}}\toprule
& \footnotesize{0} & \footnotesize{1}  & \footnotesize{2} & \footnote{3} & \footnote{4} & \footnote{5}  \\  
\midrule
\hspace{10pt}\scriptsize{Training set} & \scriptsize{$4,202,040$} & \scriptsize{$4,093,033$} &  \scriptsize{$109,217$} &  \scriptsize{$2,262$} &  \scriptsize{$9$} &  \scriptsize{$9$}  \\
\hspace{10pt}\scriptsize{Testing set} & \scriptsize{$1,052,062$} & \scriptsize{$1,021,898$} &  \scriptsize{$27,110$} &  \scriptsize{$568$} &  \scriptsize{$3$} &  \scriptsize{$0$}  \\
\bottomrule
\end{tabular}
\caption{\label{tab:schubert-stats-6}Statistics for the Schubert polynomial structure constants dataset for $n = 6$.}
\end{table*}


\subsection{Partial Orders on Lattice Paths}

This dataset contains pairs of lattice paths starting at $(0,0)$ and ending at $(n,n-1)$ that are only allowed to take one unit steps either north or east, and must stay below the line $y = \frac{n}{n-1}x$. They are thus encoded by a sequence of $1$'s (for steps east) and $0$'s (for steps north) of length $(n+1) + n = 2n+1$. Each pair of lattice paths is a covering pair in exactly one of the two partial orders, the Lagrange order or the matching order (pairs that are covers in both are few and were removed). The task is to predict which partial order a covering pair belongs to.

 Each line in a file is the concatenation of two $0$-$1$ sequences (one for each path) for a length $4n+2$ row of $0$'s and $1$'s. The lattice paths are separated by `;'. 
 
\begin{figure}[h!]
    \centering
\scalebox{-1}[1]{\rotatebox{270}{\includegraphics[scale=0.15]{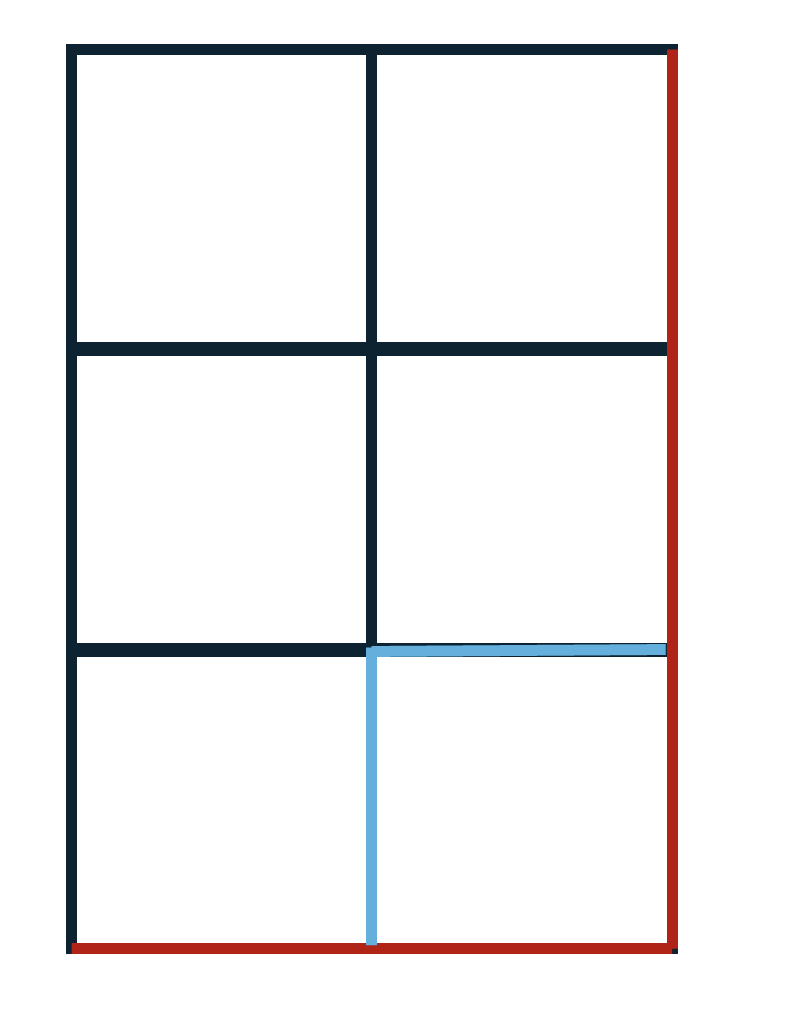}}}
    \caption{An example of two lattice paths from $(0,0)$ to $(3,2)$. These do not correspond to a cover relation.}
    \label{fig:lattice_paths}
\end{figure}

 For an $3 \times 2$ grid, the sequence:
 \begin{equation*}
     1,1,1,0,0;1,1,0,1,0
 \end{equation*}
 corresponds to the two lattice paths in Figure \ref{fig:lattice_paths}. The first is in red and second is in blue, with segments traversed by both paths colored red. 

We store files for $n = 10,11,12,13$. These are named:
\begin{itemize}
\item \begin{verbatim}lagrange_covers_test_10_9.csv \end{verbatim}
\item \begin{verbatim}lagrange_covers_test_11_10.csv \end{verbatim}
\item \begin{verbatim}lagrange_covers_test_12_11.csv \end{verbatim}
\item \begin{verbatim}lagrange_covers_test_13_12.csv \end{verbatim}
\item \begin{verbatim}lagrange_covers_train_10_9.csv \end{verbatim}
\item \begin{verbatim}lagrange_covers_train_11_10.csv \end{verbatim}
\item \begin{verbatim}lagrange_covers_train_12_11.csv \end{verbatim}
\item \begin{verbatim}lagrange_covers_train_13_12.csv \end{verbatim}
\item \begin{verbatim}matching_covers_test_10_9.csv \end{verbatim}
\item \begin{verbatim}matching_covers_test_11_10.csv \end{verbatim}
\item \begin{verbatim}matching_covers_test_12_11.csv \end{verbatim}
\item \begin{verbatim}matching_covers_test_13_12.csv \end{verbatim}
\item \begin{verbatim}matching_covers_train_10_9.csv \end{verbatim}
\item \begin{verbatim}matching_covers_train_11_10.csv \end{verbatim}
\item \begin{verbatim}matching_covers_train_12_11.csv \end{verbatim}
\item \begin{verbatim}matching_covers_train_13_12.csv \end{verbatim}
\end{itemize}

The first word (`Lagrange' or `matching') gives the label, the third word says whether it is train or test, and the final two numbers give $n$ and $n-1$. 

Sage \cite{sage} was used to compute the covering pairs in the Lagrange and matching ordering; the script used to generate the data is available on the GitHub page.
We removed instances that were covering pairs in both the Lagrange partial order and the matching partial order (this was 21, 40, and 79 instances for $n = 10, 11,$ and $12$ respectively).

The dataset statistics can be found in \cref{tab:lattice-stats-10}-\cref{tab:lattice-stats-13}.

\begin{table*}\centering
\begin{tabular}{@{}lccc@{}}\toprule
& \footnotesize{Lagrange order} & \footnotesize{Matching order}  & \footnotesize{Total} \\  
\midrule

\hspace{10pt}\scriptsize{Training set} & \scriptsize{$7,5589$} & \scriptsize{$3,875$} &  \scriptsize{$11,433$} \\
\hspace{10pt}\scriptsize{Testing set} & \scriptsize{$1,895$} & \scriptsize{$968$} &  \scriptsize{$2,863$}\\
\bottomrule
\end{tabular}
\caption{\label{tab:lattice-stats-10}Statistics for the lattice paths dataset for paths from $(0,0)$ to $(10,9)$.}
\vspace{-.3cm}
\end{table*}

\begin{table*}\centering
\begin{tabular}{@{}lccc@{}}\toprule
& \footnotesize{Lagrange order} & \footnotesize{Matching order}  & \footnotesize{Total} \\  
\midrule

\hspace{10pt}\scriptsize{Training set} & \scriptsize{$26,427$} & \scriptsize{$13,424$} &  \scriptsize{$39,851$} \\
\hspace{10pt}\scriptsize{Testing set} & \scriptsize{$6,601$} & \scriptsize{$3,355$} &  \scriptsize{$9,956$}\\
\bottomrule
\end{tabular}
\caption{\label{tab:lattice-stats-11}Statistics for the lattice paths dataset for paths from $(0,0)$ to $(11,10)$.}
\vspace{-.3cm}
\end{table*}

\begin{table*}\centering
\begin{tabular}{@{}lccc@{}}\toprule
& \footnotesize{Lagrange order} & \footnotesize{Matching order}  & \footnotesize{Total} \\  
\midrule

\hspace{10pt}\scriptsize{Training set} & \scriptsize{$93,218$} & \scriptsize{$46,976$} &  \scriptsize{$140,194$} \\
\hspace{10pt}\scriptsize{Testing set} & \scriptsize{$23,324$} & \scriptsize{$11,749$} &  \scriptsize{$35,073$}\\
\bottomrule
\end{tabular}
\caption{\label{tab:lattice-stats-12}Statistics for the lattice paths dataset for paths from $(0,0)$ to $(12,11)$.}
\vspace{-.3cm}
\end{table*}

\begin{table*}\centering
\begin{tabular}{@{}lccc@{}}\toprule
& \footnotesize{Lagrange order} & \footnotesize{Matching order}  & \footnotesize{Total} \\  
\midrule
\hspace{10pt}\scriptsize{Training set} & \scriptsize{$331,065$} & \scriptsize{$166,304$} &  \scriptsize{$497,369$} \\
\hspace{10pt}\scriptsize{Testing set} & \scriptsize{$82,789$} & \scriptsize{$41,580$} &  \scriptsize{$124,369$}\\
\bottomrule
\end{tabular}
\caption{\label{tab:lattice-stats-13}Statistics for the lattice paths dataset for paths from $(0,0)$ to $(13,12)$.}
\vspace{-.3cm}
\end{table*}

\subsection{Mutation Equivalent Quivers}

The task associated with this dataset is matching a quiver to one of several possible mutation equivalence classes. Thus, the input is a quiver with $11$ nodes encoded by its $11 \times 11$ adjacency matrix and the label is one of $7$ different equivalence classes: $A_{11},BB_{11},BD_{11},BE_{11},D_{11},DE_{11},E_{11}$. The files are organized by train and test for each of these classes. All mutation classes were generated using Sage \cite{sage}, the code is available on the GitHub page. For the quiver mutation classes that are not mutation finite, the datasets contain quivers generated up to a certain depth, which is the distance from the original quiver, measured by number of mutations. The depth is specified in the filename and was chosen to achieve as close to a balanced dataset as possible. 

The file names are:
\begin{itemize}
\item \begin{verbatim}A_11_bmatrices_test.csv \end{verbatim}
\item \begin{verbatim}A_11_bmatrices_train.csv \end{verbatim}
\item \begin{verbatim}BB_11_depth10_bmatrices_test.csv \end{verbatim}
\item \begin{verbatim}BB_11_depth10_bmatrices_train.csv \end{verbatim}
\item \begin{verbatim}BD_11_depth9_bmatrices_test.csv \end{verbatim}
\item \begin{verbatim}BD_11_depth9_bmatrices_train.csv \end{verbatim}
\item \begin{verbatim}BE_11_depth8_bmatrices_test.csv \end{verbatim}
\item \begin{verbatim}BE_11_depth8_bmatrices_train.csv \end{verbatim}
\item \begin{verbatim}D_11_bmatrices_test.csv \end{verbatim}
\item \begin{verbatim}D_11_bmatrices_train.csv \end{verbatim}
\item \begin{verbatim}DE_11_depth9_bmatrices_test.csv \end{verbatim}
\item \begin{verbatim}DE_11_depth9_bmatrices_train.csv \end{verbatim}
\item \begin{verbatim}E_11_depth9_bmatrices_test.csv \end{verbatim}
\item \begin{verbatim}E_11_depth9_bmatrices_train.csv \end{verbatim}
\end{itemize}
Within a file, each row is a flattened adjacency matrix encoded in row major order. The statistics of the dataset can be found in Table~\ref{tab:quiver-statistics}.

\begin{table*}\centering
\begin{tabular}{@{}lcccccccc@{}}\toprule
& \footnotesize{$A_{11}$} & \footnotesize{$B_{11}$}  & \footnotesize{$BD_{11}$} & \footnotesize{$BE_{11}$} & \footnotesize{$D_{11}$} & \footnotesize{$DE_{11}$} & \footnotesize{$E_{11}$} & \footnotesize{Total} \\  
\midrule

\hspace{10pt}\scriptsize{Training set} & \scriptsize{$11,940$} & \scriptsize{$27,410$} &  \scriptsize{$23,651$} &  \scriptsize{$22,615$} & \scriptsize{$25,653$} &  \scriptsize{$23,528$} &  \scriptsize{$28,998$} & \scriptsize{$163,795$}\\
\hspace{10pt}\scriptsize{Testing set} & \scriptsize{$2,984$} & \scriptsize{$6,852$} &  \scriptsize{$5,912$} &  \scriptsize{$5,653$} & \scriptsize{$6,413$} &  \scriptsize{$5,881$} &  \scriptsize{$7,249$} & \scriptsize{$409,44$}\\
\bottomrule
\end{tabular}
\caption{\label{tab:quiver-statistics}Statistics of the quiver mutation equivalence dataset.}
\vspace{-.3cm}
\end{table*}


\subsection{Weaving Patterns}

Weaving patterns of size $n \times n-1$ are a special type of matrix containing entries in $\{1,2,\dots,n\}$. They correspond to representations of the longest word permutation of $n$ elements (the permutation that sends $1 \mapsto n$, $2 \mapsto n-1$, etc.). This task involves trying to identify weaving patterns among matrices that look like weaving patterns but are not.

Each matrix is stored on a single line in row-major format. For instance,
\begin{verbatim}
(0, 1, 2, 3, 3, 2, 3, 4, 2, 3, 2, 1, 5, 4, 3, 2)
\end{verbatim}

We provide files for $n = 6,7$. The files are:
\begin{itemize}
\item \begin{verbatim}weaving_patterns_6_train.txt \end{verbatim}
\item \begin{verbatim}weaving_patterns_7_train.txt \end{verbatim}
\item \begin{verbatim}weaving_patterns_6_test.txt \end{verbatim}
\item \begin{verbatim}weaving_patterns_7_test.txt \end{verbatim}
\end{itemize}
Positive examples were generated by a program written in Java script. Negative examples were generated by loading positive examples and Python and perturbing them. Code for both of these can be found on the GitHub page.

Dataset statistics can be found in \cref{tab:weaving-stats-6} and \cref{tab:weaving-stats-7}.

\begin{table*}\centering
\begin{tabular}{@{}lcc@{}}\toprule
& \footnotesize{Weaving pattern} & \footnotesize{Non-weaving pattern}  \\  
\midrule
\hspace{10pt}\scriptsize{Training set} & \scriptsize{$634$} & \scriptsize{$1,116$}  \\
\hspace{10pt}\scriptsize{Testing set} & \scriptsize{$275$} & \scriptsize{$467$} \\
\bottomrule
\end{tabular}
\caption{\label{tab:weaving-stats-6}Statistics of the weaving pattern dataset for $n = 6$.}
\vspace{-.3cm}
\end{table*}

\begin{table*}\centering
\begin{tabular}{@{}lcc@{}}\toprule
& \footnotesize{Weaving pattern} & \footnotesize{Non-weaving pattern}  \\  
\midrule
\hspace{10pt}\scriptsize{Training set} & \scriptsize{$17,388$} & \scriptsize{$96,012$}  \\
\hspace{10pt}\scriptsize{Testing set} & \scriptsize{$7,310$} & \scriptsize{$41,290$} \\
\bottomrule
\end{tabular}
\caption{\label{tab:weaving-stats-7}Statistics of the weaving pattern dataset for $n = 7$.}
\vspace{-.3cm}
\vspace{-.3cm}
\end{table*}

\section{Model Performance}
\label{appendix:model_performance}

Amongst all our small models, we found that MLPs tended to perform most consistently across tasks (at least given our simple training set up). Transformers also performed well but struggled in a few cases (such as the lattice paths dataset). This could be due to a number of factors including a hyperparameter optimization set-up that is non-optimal for transformers or the need for larger transformers. We also see the sensitivity bias of transformers \cite{hahn2024sensitive} as being counter to many problems in combinatorics where the ground truth label can change radically with small changes to the input (at least for many standard representations of combinatorial gadgets). A toy example of this is the parity of a permutation which can change with a single digit to the one-line notation representation of the permutation. Developing novel representations of input data that result in the task label no longer being sensitive to small changes would be an interesting direction of research, but seems challenging in cases where one does not a priori know how to solve a problem.

Unsurprisingly, we found that larger datasets were generally associated with better model performance. This is true even in the case where generating a larger dataset required increasing $n$, and thus making the problem potentially more complex (e.g., working with partitions of $n+1$ rather than partitions of $n$). There were some tasks however that seem hard even when the dataset size is increased. As can be seen in Table \ref{tab:baselines_regression}, performance regressing symmetric group characters is very poor. This may relate to the complexity of the task (calculating symmetric group characters is known to belong to $\#P$ \cite{hepler1996complexity}). It may also relate to the distribution of symmetric group characters which has a very long tail.

\subsection{Small Model Hyperparameters}
\label{appendix:baseline_hyperparameters}

For our baselines, we train encoder-only transformer models, standard feedforward multi-layer perceptron (MLP) models with ReLU non-linearities, and logistic regression on the classification tasks and the same architectures of transformers and MLPs along with linear regression for the regression tasks. 
\begin{itemize}
    \item To optimize MLP models we performed a simple grid search across 
    \begin{itemize}
        \item learning rates ($0.001$, $0.0005$, and $0.0001$), 
        \item depths ($1$, $2$, $3$, and $4$), 
        \item and constant hidden dimension ($32$, $64$, $128$, and $256$).
    \end{itemize}
    \item To optimize the transformer models we performed a similar grid search but with hyperparameters
    \begin{itemize}
        \item learning rates ($0.001$, $0.0005$, and $0.0001$), 
        \item model dimensionality ($20$, $40$, and $80$), 
        \item depths ($2$ and $4$), 
        \item and number of heads ($4$, $6$, and $8$). 
    \end{itemize}
\end{itemize}           
All optimization was performed in Pytorch with the Adam optimizer on one Nvidia $A100$. After finding an optimal hyperparameter setting, we trained $3$ models, averaged their performance, and generated $95\%$ confidence intervals to measure the variability in performance with these hyperparameters.

Linear and logistic regression was performed with sklearn using standard hyperparameters.

\subsection{LLM Evaluation Procedure}
\label{appendix:llm_hyperparameters}

We evaluated Claude 3.5 Sonnet, GPT-4o Mini, GPT-4o, and o1-mini on the mHeight and Schubert polynomials datasets using a simple in-context learning setup and a simple program synthesis setup. When using in-context learning, we tried using $0$, $10$, $50$, and $100$ examples and also tested using chain-of-thought prompting. Finally, we explored providing the model with some background on the problem. The best setting depended on the model and task. An example template for the in-context learning prompt when chain-of-thought is used is:
\begin{verbatim}
"You are tasked with solving a classification problem. 
Here is high-level information about the dataset:\n{dataset_info}\n\n" 
+ f"{few_shot_str}" + "Before answering with your Python code, 
reason in a step-by-step manner as to get the right answer.\n\n"
\end{verbatim}
where `dataset\_info' is a description of the dataset and task and `few\_shot\_str' are the few-shot examples. When chain-of-thought reasoning is not used the last component is changed to 
\begin{verbatim}
"Do not provide any additional reasoning or explanation. 
Just provide your answer at the end on its own line in the form 'ANSWER: 
$ANSWER' (without quotes) where $ANSWER is the answer to the question."
\end{verbatim}

In the code synthesis version of the experiments, models were asked to write a Python program that solves the task using only Sage \cite{sage}, Numpy, and SymPy. No model ended up using either Sage or SymPy. The same prompt was used to generate 100 examples. The best program was chosen by evaluating each on the test set.
\begin{verbatim}
Your job is to write a Python function that solves the classification problem. 
You will be given some examples of a classification problem from the 
'{dataset}' dataset.

Write a function 'predict' that takes an input in a Python list and returns 
an integer  as the classification result.

Here is information about the dataset:
{dataset_info}
 
Avoid using machine learning or model calls; rather, embed the logic in 
Python code.
Rather than use shallow pattern matching or using simple patterns, try
to analyze the underlying combinatorial logic of the examples. Note 
that the datagenerating process for this dataset is a combinatorial algorithm.
You may want to use numpy and sympy for math operations or sage for 
cominatorics, however this is optional. If you do use them, *make sure 
to import them within your function*.
 
Below are a few examples from the training set:
{training_examples}
 
{instructions}
Your final answer should be valid Python code enclosed in triple 
backticks. This program will be evaluated on the test set.
\end{verbatim}

All of our experiments with LLMs used AI Inspect \cite{UK_AI_Security_Institute_Inspect_AI_Framework_2024}.



\end{document}